# Constructing a norm for children's scientific drawing: Distribution features based on semantic similarity of large language models

Yi Zhang[1,#], Fan Wei[2,#], Jingyi Li[3,#], Yan Wang[1], Yanyan Yu[1], Jianli Chen[1], Zipo Cai[1], Xinyu Liu[1], Wei Wang[1], Sensen Yao[4], Peng Wang[5] and Zhong Wang[6,*]

[1]First Primary School, Beijing 100071, China
[2]Fangzhuang Primary School, Beijing 100078, China
[3]College of Elementary Education, Capital Normal University, Beijing 100048, China
[4]Beijing Fengtai Institute of Education, Beijing 100073, China
[5]Hepingli No.9 Primary School, Beijing 100013, China
[6]Institute of Clinical Education, Beijing Doers Education Consulting Co., Ltd, Beijing 100027, China

*Corresponding author: Institute of Clinical Education, Beijing Doers Education Consulting Co., Ltd, Dongzhimenwai St, 36, Dongcheng, Beijing 100027, China. E-mail: wangzhong@bdice.ac.cn.

[#]Yi Zhang, Fan Wei, and Jingyi Li are the co-first authors.

## Abstract

The use of children's drawings to examining their conceptual understanding has been proven to be an effective method, but there are two major problems with previous research: (i) The content of the drawings heavily relies on the task, and the ecological validity of the conclusions is low. (ii) The interpretation of drawings relies too much on the subjective feelings of the researchers. To address this issue, this study uses the Large Language Model (LLM) to identify 1420 children's scientific drawings (covering nine scientific themes/concepts) and uses the word2vec algorithm to calculate their semantic similarity. The study explores whether there are consistent drawing representations for children on the same theme and attempts to establish a norm for children's scientific drawings, providing a baseline reference for follow-up children's drawing research. The results show that the representation of most drawings has consistency, manifested as most semantic similarity >0.8. At the same time, it was found that the consistency of the representation is independent of the accuracy (of LLM's recognition), indicating the existence of consistency bias. In the subsequent exploration of influencing factors, we used Kendall rank correlation coefficient to investigate the effects of "sample size," "abstract degree," and "focus points" on drawings and used word frequency statistics to explore whether children represented abstract themes/concepts by reproducing what was taught in class. It was found that accuracy (of LLM's recognition) is the most sensitive indicator, and data such as sample size and semantic similarity are related to it. The consistency between classroom experiments and teaching purpose is also an important factor, many students focus more on the experiments themselves rather than what they explain. In addition, most children tend to use examples they have seen in class to represent more abstract themes/concepts, indicating that they may need concrete examples to understand abstract things.

## Introduction

Children's drawing is a commonly used way of examining children's cognitive and emotional status [1]. The reason for this is not only because drawing is a pleasurable activity for children [2], but also because its effectiveness in examining children's psychological activities is considered reliable and has been extensively studied in many fields. For example, from the perspective of individual learning, drawing is used to examine children's cognition of a certain scientific concept. Brechet *et al.* [3] had over 250 French children draw their brains to understand how the "black box" of the brain works. In terms of emotions and interpersonal relationships, campus bullying is a topic of great concern to psychologists, so drawing has also been used in related research. For example, Marengo *et al.* [4] analyzed more than 600 drawings of Italian children to understand this issue.

However, there are still some fundamental issues that have not been overcome in the current psychological analysis of children's drawings, which mainly focus on methodological perspectives. This is mainly reflected in two aspects: firstly, the issue of ecological validity. As drawing is a non-verbal representation, it must be digitized through effective encoding in order to be analyzed using statistical methods. Therefore, encoding method is currently the mainstream way to study children's drawing. For example, Tolsberg *et al.* [5] used coding to understand the longitudinal development of math word problems among Estonian elementary school students. The coding rule for that study is as follows: those without drawings are coded as 9, those depicted as pictorial are coded as 1, and those depicted as schematic are coded as 2. The main difference between 1 and 2 is whether the relationship between the things is revealed. Therefore, how to develop coding rules is a crucial issue. This issue is particularly prominent in learning science, as emotions and interpersonal relationships have a large number of scales or tools (such as bullying problems [6]), and some drawing studies on this issue can use these general tools [4]. However, many researchers in learning science area had to adopt self-developed coding rules. In a psychology journal, there were a total of 12 original studies on the topic of children's drawing research, of which only four studies did not involve self-developed coding at all (only 33.3%) [7].

The high proportion of self-developed coding naturally raises a question: how can these highly task-dependent coding rules be transferred and generalized to other drawing studies? In other words, different studies have different tasks, so how should other drawing studies refer to the results of these codes? Obviously, existing drawing research lacks a unified norm, which allows students to draw a certain field or object relatively freely under the condition of no tasks or fewer tasks, in order to fully expose their drawing preferences for that object in the naturally state. If this norm exists, different drawing tasks can be compared with it, then there is a common reference frame for drawing research on the same theme.

The second issue is more obvious: the understanding of drawing is one of the most prominent areas of individual heterogeneity, and different people's perspectives on the same drawing may be completely different [8]. Therefore, for studies do not utilize machines, consistency testing is an essential step—the same drawing must be identified by at least two researchers, and the results of these two identifications must pass consistency testing (such as Cohen Kappa coefficient [3, 5]). On the contrary, if machine learning models are used to identify drawings, this process is likely to be omitted. Because the same model is based on the same algorithm, its generation results should be more consistent than those of researchers.

Based on the above two issues, this study aims to construct a norm of drawings for children's understanding of scientific themes and to use the Large Language Model (LLM) to understand and identify children's drawings, attempting to provide a reference frame for future research on children's conceptual understanding of drawing. This study focuses on two questions: (i) Do children's drawings have a consistent representation for the same drawing theme (such as whether the concept of solar eclipse is mostly expressed through light path diagrams)? (ii) What factors are related to the consistent representations (if it exists)?

## Materials and Methods

### Participants

Select a total of 1473 students from two primary schools in Beijing. Among them, there are 74 students in grade 4, and the rest are in grades 5 and 6. The male to female ratio is 1.01:1. The textbooks involved two versions [9, 10].

We excluded drawings that were unclear, incomplete, or clearly did not belong to the prescribed content, as well as drawings with inconsistent recognition results by LLM multiple times (see section "LLM Identification of drawings" of this article) and finally included a total of 1420 drawings in the sample.

All children's personal information is strictly confidential, and the drawings were sealed after being photographed to ensure that personal information of students were not leaked. The students' names/IDs were also removed. The method of removing is as follows: firstly, open a children's drawing using Photoshop. Secondly, select the "Clone Stamp Tool" function in the left tab and adjust the size of the icon to easily cover the name/ID. Thirdly, hold down the Alt key while selecting the area without any graphic content around it. The main purpose of this step is to let the software know which area of the image to use to overlay the image containing names/IDs. Fourth, release the Alt key and draw the area where the name/ID is located. In this way, the software can use the selected area to overlay the image of the area where the name/ID is located, achieving the goal of overlaying the name/ID. And after this operation, the screen appears natural and will not attract the attention of LLM due to missing parts of the screen, causing it to misjudge. Informed consent from the participants' guardian was obtained. This research was approved by the Research Ethical Review Committee of Beijing Doers Educational Consulting Co., Ltd on 13 March 2024 (ECS-008).

### Proceeding

#### *Collection of drawings*

Firstly, all children draw in class, i.e. they draw whatever they just learned (following the order of textbooks), without deliberately setting contents or tasks. Secondly, to ensure the consistency and interpretability of the data, all participating teachers in this study are required to strictly follow the textbook and strictly prohibit free design teaching scaffolds and guidance to prevent individual differences among teachers from interfering with the results. Thirdly, the prompt was: "We have just learned the knowledge of …, please use drawing to represent it." Other than that, no hints will be given to prevent any suggestion or misleading to students. Fourth, ensure that all students can complete the drawing.

#### *Drawings preprocessing*

Firstly, number and take photos of all the drawings. Secondly, preprocessing the photos includes two parts: (i) removing students' personal information, such as names and student IDs, and (ii) removing the theme/concept names (some students have the habit of writing the names of the themes/concepts on their drawings) to prevent prompting LLM. The solution is to use the "Clone Stamp Tool" function of Photoshop software (v.2018) to cover these names with surrounding images.

#### *LLM Identification of drawings*

Firstly, input the photos into LLM for identification. The prompt is "What is the specific content of this drawing? How did you see it from the picture?" and copy all the results and reasons generated by LLM into an Excel file for analysis.

In addition, to ensure the stability of LLM recognition, all drawings are independently recognized twice by LLM. Drawings with significantly different recognition results (such as the first recognition as a solar eclipse and the second recognition as a lunar phase) will undergo a third independent identification, and those with two occurrences of the three results will be considered as the result. If three different results occur, the drawing will be removed. The so-called "independence" refers to deleting the current page after one recognition is completed and re-recognizing it. Since the LLM used does not capture conversations with front-end users and place them in its pre training set, the above operation method can ensure the independence of the two recognitions.

This study used ChatGLM-4.0 to identify drawings. The reason for adopting this LLM is that its image recognition models mainly use convolutional neural networks (CNN) and recurrent neural networks (RNN) (derived from the self-introduction of this LLM), and these two models have been proven to be efficient and reliable in image recognition (because the two models are so famous and there are too numerous related studies, so this article will not list the literatures here). And there have been some studies published using this LLM for image recognition [11, 12], further proving its reliability in image recognition. And its language generation method adopts a model based on the transformer algorithm, which has high efficiency and reliability in language generation. More importantly, the basic version of this LLM is open to individual users for free, making it more suitable for numerous primary school teachers to identify children's drawings.

To verify that LLM's image recognition reasonings can indeed explain children's drawings representations, we counted the number of objects appearing in the drawings in all LLM image recognition reasoning (as long as one object appears in the drawings in these sentences, it is considered). The results showed that 77.4% (weighted average) of the reasoning statements for image recognition contained at least one object appearing in the drawings. This indicates that LLM is indeed describing the specific content of the drawings, rather than fabricating reasons. At the same time, in order to prevent LLM from generating image recognition reasons by reading text on the drawings, we randomly selected 16 drawings without any prompts (2 drawings for each theme/concept, and all drawings of the "solar system" concept have prompts, so this concept cannot be selected). The results showed that LLM described at least one object in all the 16 drawings.

## Semantic similarity analysis

### *Algorithm*

This study used the semi-supervised Word2vec algorithm to generate word vectors and then used the cosine similarity algorithm to calculate semantic similarity values.

The Word2vec is a deep learning algorithm for word embedding proposed by the Tomas Mikolov team at Google. Traditional spatial distance–based algorithms (such as Euclidean distance or Manhattan distance) in handling sentences with the same semantics but different representations usually have poor performance [13]. In contrast, the Word2vec algorithm creatively proposes a negative sampling method, which calculates the conditional probability of the sampled sample coming from the corpus through context word frequency statistics and then determines its polarity [14]. This method can better identify similar semantic sentences with different representations.

Cosine similarity is a widely used algorithm for calculating semantic similarity. Its main principle is to model the text as a term vector and calculate the cosine value between the term vectors to determine the degree of similarity between two texts [15]. The cosine similarity algorithm has significant advantages in semantic similarity calculation, with evidence showing that it has higher resolution than Jacobian similarity [16].

By incorporating supervision into unsupervised algorithms, it has been widely applied in NLP semantic analysis and comparative research on cross-class text learning between machines and humans, demonstrating its advantages [17].

### *Proceeding of semantic similarity analysis*

Firstly, we collated all the statements generated by LLM for image recognition, match the statements with the image numbers one by one, and outputted the results as an Excel file. Then, used the JIEBA tokenizer to segment the statements in this Excel file to words (removed all punctuation marks). Then, we trained a semi-supervised Word2vec algorithm (dimension 100, context window size 5, negative sampling 5) to embed word vectors into the segmentation results and then used the word vector embedding results to calculate the average word vector of each sentence. Finally, the cosine similarity algorithm was used to calculate the semantic similarity between all pairs of sentences, thereby forming a similarity matrix.

In addition, although the theoretical range of semantic similarity calculated by the cosine similarity algorithm should be [−1,1], all semantic similarity results in this article are greater than 0. Therefore, for the sake of simplicity in calculation, all calculations regarding the concentration of semantic similarity in the following text were defaulted the value range of [0,1] instead of [−1,1].

### *The semi-supervised algorithms*

Due to Word2vec being an unsupervised algorithm, supervised learning is clearly superior in most cases in NLP processing. Therefore, we introduce keywords to transform the above algorithm into a semi-supervised learning method, thereby increasing the accuracy of word vector embedding.

The specific method was to first use the JIEBA tokenizer to segment the text to be analyzed and remove all punctuation marks. Second, the TF-IDF algorithm to search for keywords in the text was used and assigned weights to these keywords. The number of keywords of each text was set at 10, because after repeated experiments, when the number exceeds 10 (such as 15), there will be a large number of high-frequency function words (such as "this" and "yes") in the keywords. Third, semi-supervision by adding Python statements was implemented. Finally, the generated word vectors into an Excel file to calculate the average word vector for each sentence in the original text was embedded.

## Distribution analysis and visualization

Used a heat map to represent the similarity values of the semantic similarity matrix for each theme/concept and used a histogram to represent the specific distribution of similarity in [0,1].

Specifically, as this article mainly focuses on the concentration of semantic similarity in LLM image recognition reasons. But the traditional quartiles from Q1 to Q3 can only reflect the concentration of 50% of the data. In order to intuitively reflect the degree of concentration of more data, this article constructs a new indicator based on the construction method of quartiles: all semantic similarity values of a concept are arranged in ascending order and evenly divided into eight parts. In this way, from the Q1 (equivalent to the Q1 of quartiles) to Q7 (equivalent to the Q3 of quartiles) encompass three-quarters (75%) of all data. So, calculating the concentration from Q1 to Q7 of the new indicator is equivalent to knowing how concentrated 75% of the total data is (locally referred to as the "75% Region"). The smaller the value, the higher the numerical concentration, which is used to examine whether the theme/concept has a consistent representation.

## Exploration of related factors

Finally, we used Kendall rank correlation coefficient (Kendall's $\tau$) to calculate the effects of three factors: Sample Size, Abstract Degree of Themes/Concepts, and Reasoning Degree of Themes/Concepts on the Accuracy, Semantic Similarity, and 75% Region of children's drawing representations.

There are two reasons for using this indicator (Kendall's τ): firstly, this article is only an observational study, not a control study, so exploring correlation is more suitable than establish the causal relationship [18]. Secondly, the related factor exploration in this article aims to understand the impact of a certain factor on nine concepts, which means we only have a sample size of 9. The commonly used Pearson correlation coefficient is an indicator that strictly depends on sample size, but due to the limitations of researchers, we do not have the ability to explore more concepts of children's drawings. The Kendall's τ is more suitable for dealing with low sample size correlation problems, so we chose this indicator to explore the relevant factors.

At the same time, we also examined the reproduction rate of children's experiments/phenomena taught in class (mainly for abstract themes/concepts) through the method of word frequency. Because we suspect that students may have difficulty grasping the connotations of abstract concepts, they may use experiments/phenomena taught in class to represent these concepts in drawings.

# Results

## Overall

The total number of samples is 1420, covering nine scientific themes or concepts: Circuits, Solar eclipses, Boiling, Solar system, Increasing the carrying capacity of a boat, Life history of a plant, Buoyancy, Electromagnetics, and Physical and chemical changes. The average recognition accuracy of LLM is 46.7%, and the median semantic similarity is 0.9885. In addition, the mean of the "75% Region" is 0.0035 (i.e. occupying 0.35% of [0,1]). The data is detailed in Table 1.

From the overall situation, it can be seen that LLM has a low recognition accuracy for image drawing content, but the semantic similarity concentration is high. This seems to imply that there was a consistency preference in the representation of most of the drawings, but there is a possibility of "collective bias" in this preference (i.e. those children's drawing representations were inaccurate, but the way these erroneous representations were consistent).

## Results of each theme/concept

### Circuit

The main learning content (experiment) of this concept is to light up a bulb to let students know that the circuit must be closed in order to work (i.e. form a closed loop).

The average accuracy of LLM image recognition is 84.3%, which is the highest among all nine themes/concepts, indicating that most students can grasp the correct representation method of this concept.

**Table 1.** Overall situation of the samples (where 75% refers to the 75% Region, with smaller values being more concentrated).

| Concepts | Sample size | Accuracy | Median value | 75% |
|---|---|---|---|---|
| **Circuit** | 267 | 0.843 | 0.9998 | 0.02 |
| **Solar eclipses** | 214 | 0.44 | 0.9999 | 0 |
| **Boiling** | 48 | 0 | 0.967 | 1.49 |
| **Solar system** | 201 | 0.65 | 0.9999 | 0 |
| **Increasing the carrying capacity of a boat** | 52 | 0.033 | 0.987 | 1.29 |
| **Life history of a plant** | 368 | 0.712 | 0.9999 | 0.01 |
| **Buoyancy** | 40 | 0.5 | 0.944 | 3.38 |
| **Electromagnetics** | 70 | 0.514 | 0.9995 | 0.12 |
| **Physical and chemical changes** | 160 | 0.513 | 0.9999 | 0 |

The median value of semantic similarity is 0.9998, and 75% Region is concentrated within 0.02% of [0,1], indicating that the vast majority of students have a high degree of consistency in the representation of the form of circuits. Considering the high accuracy, it indicates that the majority of students' drawings are correct and consistent. For specific details, please refer to Figs 1 and 2, where Fig. 1 is a numerical analysis and Fig. 2 is the "representative drawings." The so-called "representative drawings" refer to a pair of drawings with semantic similarity located in the median of the 75% Region. As they are in the median, they should be the most representative of all drawings, and the same applies in the following.

### Solar eclipse

The main content of this concept is to understand the phenomenon and causes of solar eclipses, but it does not require representation using light paths.

The average accuracy of LLM image recognition is 44%, which is basically at the average level of nine themes/concepts.

The median value of semantic similarity is 0.9999, and 75% Region is concentrated within the 0.001% range of [0,1], indicating that the vast majority of students have a high degree of consistency in the representation of this concept (Figs 3 and 4). However, it is worth noting that the accuracy of LLM image recognition is not high, indicating the possibility of "collective bias" in students' representation. We found a possible reason for the high consistency rate but low accuracy rate when rechecking children's drawings: some students did not annotate any explanation (only drew three circles representing the sun, earth, and moon, without any signs), and the drawings did not look like them, which led LLM to mistake them for bicycle chains or other unrelated objects.

### Boiling

The main content (experiment) of this concept is to boil water through alcohol lamp. The textbook requires students to know the phenomenon of water boiling and the temperature changes during this process (i.e. it keeps rising until about 100°C and no longer changes).

The average accuracy of LLM's image recognition is 0, ranking at the bottom of the nine experiments, indicating that all students' representations misled LLM.

The median value of semantic similarity is 0.967, and 75% Region is concentrated within the 1.49% range of [0,1]. Although the 75% Region belongs to the second to last of the nine concepts, the concentration is still very high, indicating that the vast majority of students have a high degree of consistency in the representation of boiling forms (Fig. 5). However, the image recognition results of LLM are completely unrelated to this concept, indicating that almost all student representations have problems. This result is very surprising, as can be seen from the representative drawings in Fig. 6 where the problem lies: the vast majority of students' drawings focus on the experimental process and equipment, with almost no description of the changes in water during this process. This experiment does indeed require the use of this experimental equipment, but these devices are not what this class is going to learn, which means that students' focus is entirely on the experimental activity itself—rather than what the experiment is going to explain (changes in water), which is truly unexpecting.

It should be noted that we modified a keyword while tuning the model training related to this concept. Due to the significant difference between the results provided by the TF-IDF and our feeling, we conducted a retrospective analysis of all processes and found that an important word, "chemistry," was not recognized as a keyword by TF-IDF. However, this word appeared very frequently in the reasons given by LLM (about 49% of reason sentences mentioned this word). Since the TF-IDF algorithm is a statistical method based on word frequency, it cannot truly understand the meaning of the text, and the word "chemistry" appears only once in each sentence, far less frequently than "this," so it is not surprising that it is excluded by algorithm. Based on this, we selected the function word with the lowest weight from the 10 keywords provided by TF-IDF, replaced it with "chemistry," modified the corresponding word weight, and attempted to retrain. The results were very close to our intuitive feeling.

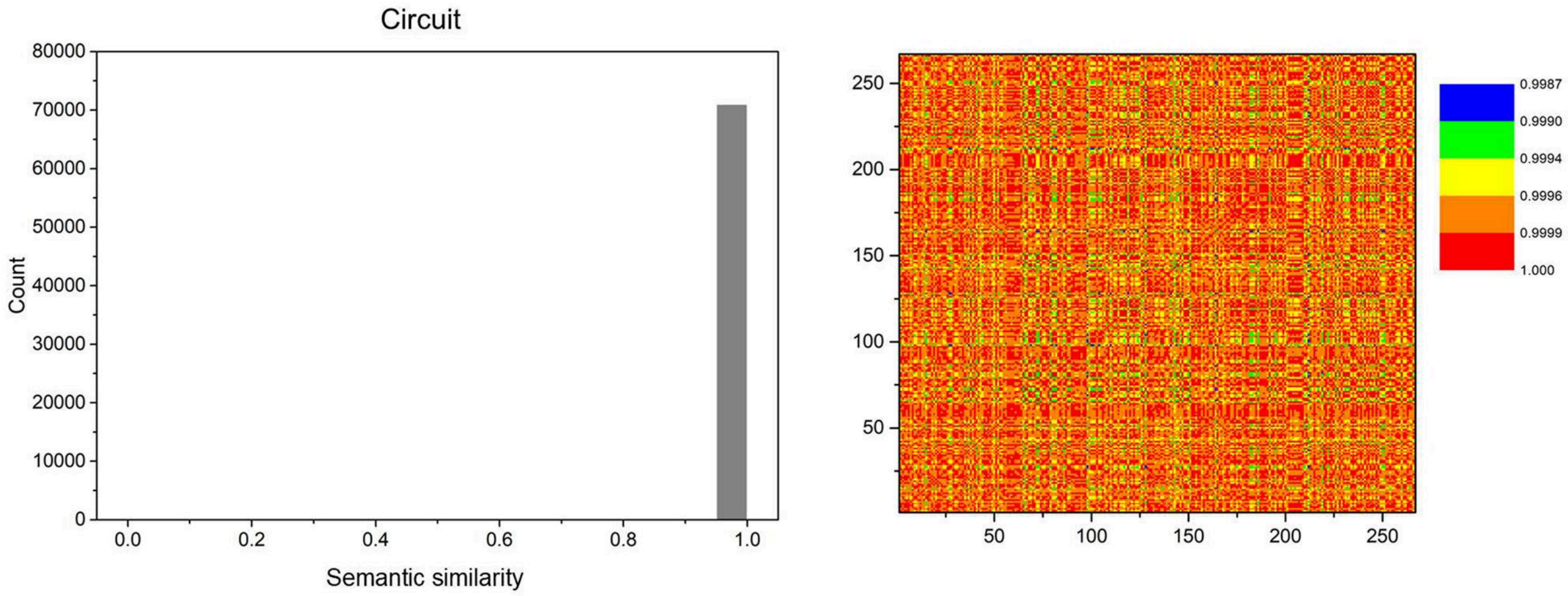

**Figure 1** Distribution diagram (left) and heatmap (right) of semantic similarity of the concept of "Circuit." The heatmap on the right specifically refers to the semantic similarity data distribution of the ≥0.9 part (the same below). The horizontal axis of the distribution chart represents the distribution of semantic similarity between concepts [0,1], and the vertical axis "Count" represents the total amount of semantic similarity data contained in the concept similarity matrix. The horizontal and vertical axes of the heatmap represent the sample size of the concept (since each image corresponds to a recognition reason, it also represents the number of recognition reasons), so the heatmap represents the semantic similarity between any two recognition reasons of the concept. In terms of color, the semantic similarity between all [0.9, 1] is divided into five levels based on the distribution of semantic similarity of the concept, represented by blue, green, yellow, orange, and red colors from low to high (excluding [0.9998, 1] and [0.9999, 1])

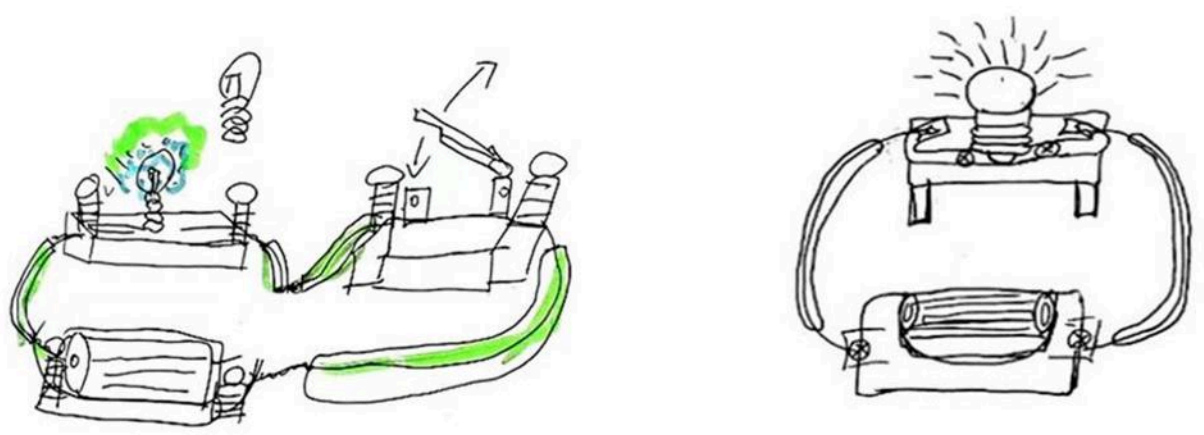

**Figure 2** Representative drawings of the concept of "Circuit"

### *Solar system*

The main content of this concept is to understand the solar system, especially the Sun and the eight major planets and their relative positions, but it does not require students learn the motion state and trajectory.

The average accuracy of LLM image recognition is 0.65, which is higher than the median value, indicating that most students can correctly represent this concept.

The median value of semantic similarity is 0.9999, and 75% Region is concentrated within the 0.001% range of [0,1]. Similar to the concept "Circuits," high accuracy and consistency in representation indicate that children have a correct and consistent understanding of this concept (Figs 7 and 8).

### *Increasing the carrying capacity of a boat*

The main content of this theme in textbook is understanding the relationship between buoyancy and the volume of displacement by changing the volume and load capacity of the boat. But the presentation of this lesson in the textbook is quite unique: first, boats with different bottom areas were made using aluminum foil of the same area, and then small iron hoops were added to different boats to increase their weight and test their load-bearing capacity. Ultimately, this guides students to identify factors associated with buoyancy. That is to say, students cannot directly adjust variables, but need to first create a "boat."

The average accuracy of LLM's image recognition is 0.033, second only to "boiling" and ranking second to last among the nine concepts, indicating that the vast majority of students' representations have misled LLM.

The median value of semantic similarity is 0.987, and 75% Region is concentrated within the 1.29% range of [0,1]. Overall, the result of this concept is very similar to "Boiling": with low accuracy but high semantic concentration, indicating that children's drawing misled LLM in the same representation way (Fig. 9).

However, the reason for this result differs greatly from the concept of "Boiling," which is due to students' excessive focus on the experimental process rather than the experimental purpose, and the problems with this theme are more caused by the textbook. As mentioned above, the original goal of this lesson was to understand the relationship between buoyancy and the volume of displacement. However, in terms of presentation, the textbook first requires students to use aluminum foil to make a boat and then complete the task of adjusting the weight. That is to say, there is a "plot transition" in this lesson, and from the representative drawings below, it can be seen that the vast majority of children were stuck at this transition point—they were most impressed with how to make a boat, rather than adjusting variables. Therefore, the most frequently depicted object in the drawing was a square aluminum foil paper with creases left by the boat making around it (Fig. 10). This directly led to errors in the overall direction of LLM: from the perspective of word frequency statistics, 59.6% of LLM's image recognition reasons showed that it was a square box–like object that had nothing to do with the boat.

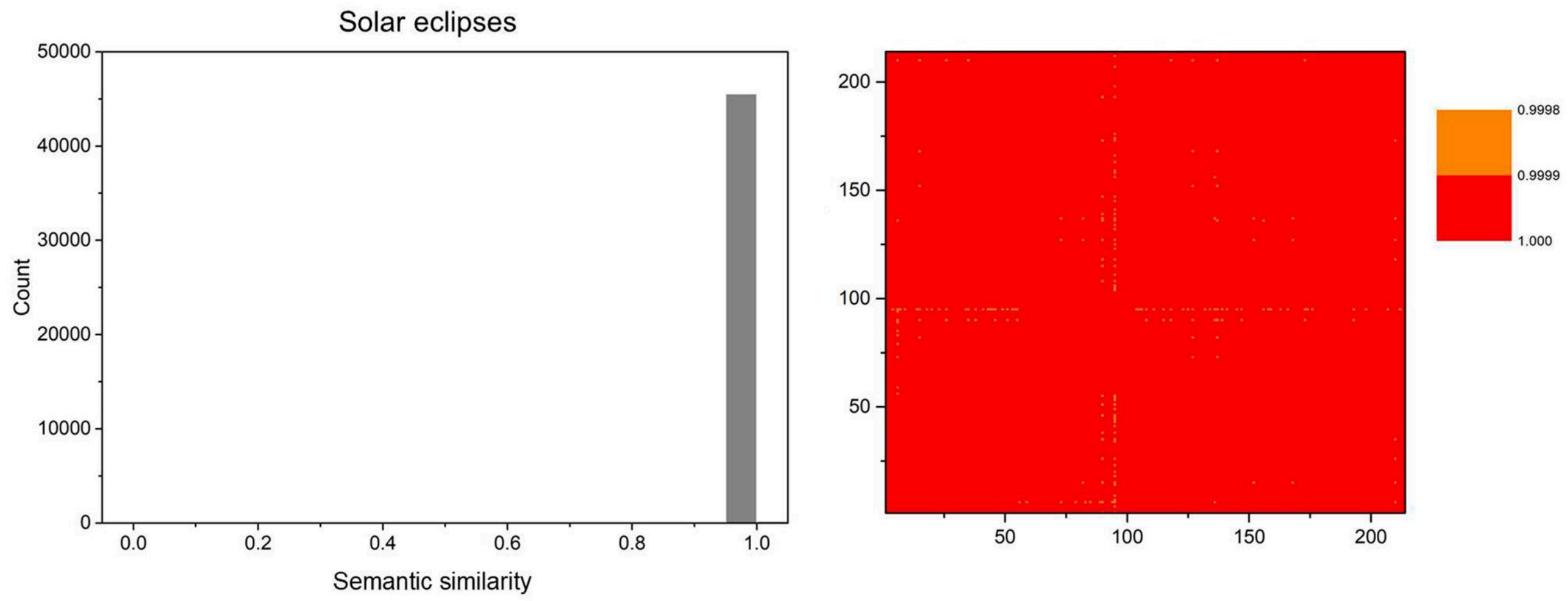

**Figure 3** Distribution diagram (left) and heatmap (right) of semantic similarity of the concept of "Solar eclipse"

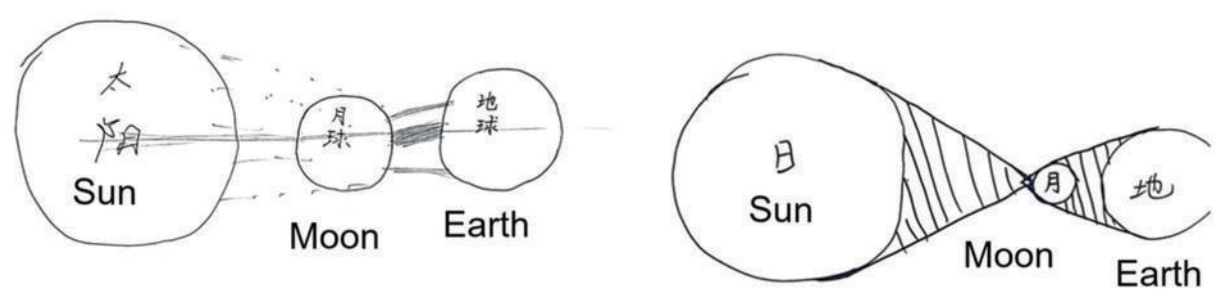

**Figure 4** Representative drawings of the concept of "Solar eclipse"

### *Life history of a plant*

The main content of this theme in the textbook is the life history of a plant, which refers to the entire process of the plant from seeds to flowering and fruiting. In the previous classes, students have already gone through tasks such as cultivating and caring for plants and recorded the growth and changes of the plants. Therefore, the main purpose of this lesson is to establish a comprehensive impression of the entire process of the plant development and changes through previous records.

The average accuracy of LLM image recognition is 0.712, ranking second among the six concepts, indicating that most students can correctly represent this theme.

The median value of semantic similarity is 0.9999, and 75% Region is concentrated within the 0.001% range of [0,1]. Figures 11 and 12 explain that children have a correct and consistent understanding of this concept.

### *Buoyancy*

This theme is very similar to the goal of "increasing the carrying capacity of a boat," which is to control the sinking or floating of the boat by changing its weight and shape. These two themes come from two different versions of textbooks, so there is some overlap in content. But unlike "increasing the carrying capacity of a boat," this lesson did not ask students to make a boat, but instead posed the core question at the beginning: how to keep a small boat floating on the water sink? Therefore, when learning this theme, the entire class was focused on exploring how to adjust variables and control the buoyancy of the boat, not to make a boat. So, as will be seen in the following text, there is a significant difference between the results than "Increasing the carrying capacity of the boat."

The average accuracy of LLM's image recognition is 0.5, ranking six among nine concepts, indicating that a considerable number of students' drawings misled LLM, but it is significantly better than "Increasing the carrying capacity of the boat."

The median value of semantic similarity is 0.944, with 75% Region concentrated within [0,1] of 3.38%, ranking last among all nine concepts (Fig. 13). After rechecking all the drawings, we found the main reason for the divergence of semantic similarity: the textbook of this lesson did not strictly define what is a "boat," so as shown in the representative drawings below (Fig. 14), some students used foam plastic and leather balls. We also found that materials such as plastic balls, wood blocks, iron boxes, well-made submarine models and warship models were used in some drawings. It is precisely because the textbook does not impose too many restrictions that students exhibited rich imagination, which also leads to the diversity of LLM's descriptions of the reasons for image recognition.

### *Electromagnetics*

This concept in textbook is mainly reproduced through the experiment of electromagnetic phenomenon discovered by Danish scientist Hans C. Oersted in 1820. And enhance electromagnetic phenomena through wire winding to make coils, to explore related factors that affect the strength of this phenomenon.4

The average accuracy of LLM image recognition is 0.514, ranking fourth among nine concepts.

The median value of semantic similarity is 0.9995, and 75% Region is concentrated within the 0.11% range of [0,1], indicating that students have strong consistency in their representation of this concept (Figs 15 and 16).

### *Physical and chemical changes*

The main content of this theme is to explore various changes in the kitchen, and to understand the difference between "physical changes" and "chemical changes" based on whether new substances are produced. This theme mainly corresponds to the "Next Generation Science Education Standards" (NGSS): "When two or more different substances are mixed, a new substance with different properties may be formed" (5-PS1-4) [19].

The average accuracy of LLM image recognition is 0.513, ranking fifth among the nine concepts, i.e. in the middle.

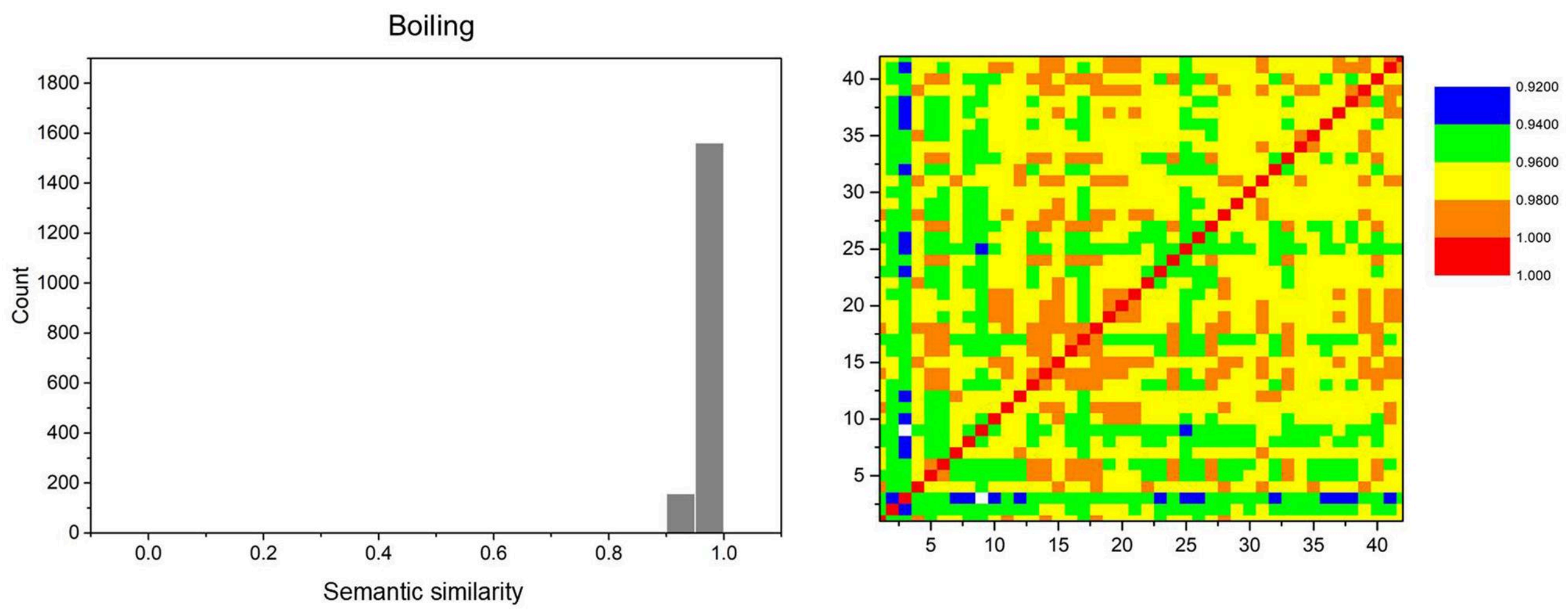

**Figure 5** Distribution diagram (left) and heatmap (right) of semantic similarity of the concept of "Boiling"

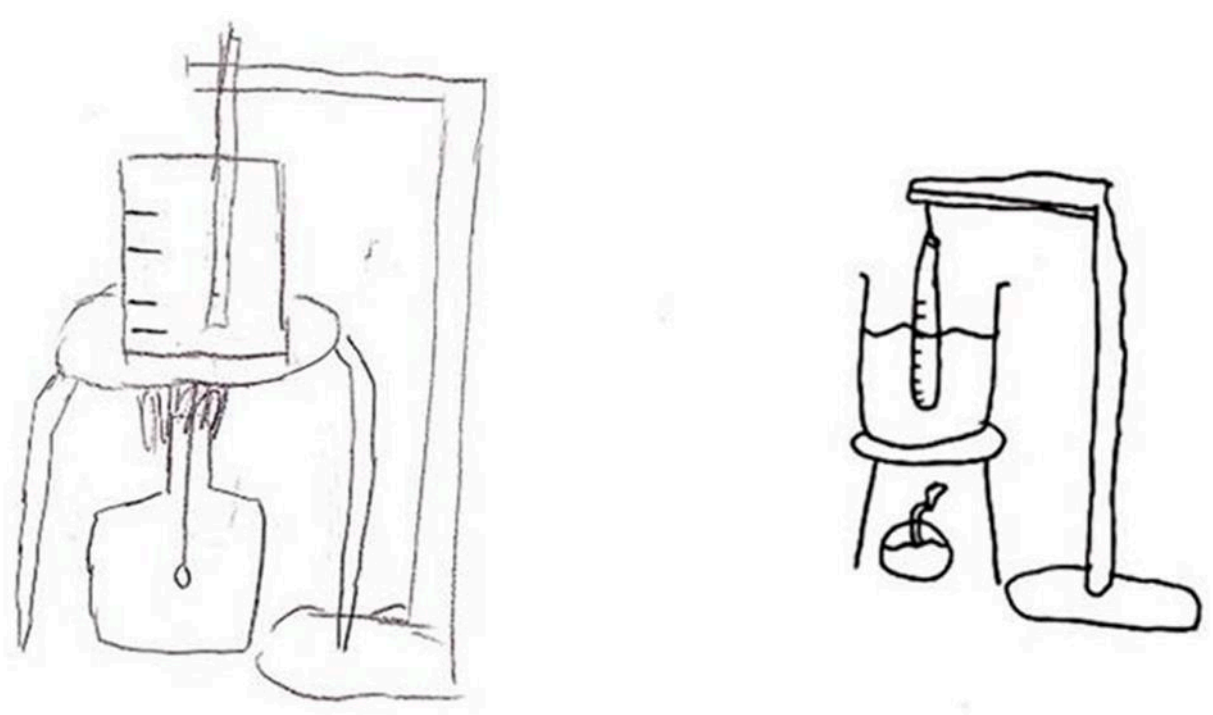

**Figure 6** Representative drawings of the concept of "boiling"

The median value of semantic similarity is greater than 0.9999, and 75% Region is concentrated within almost 0% range of [0,1], indicating that students have a huge strong consistency in their representation of this theme (Fig. 17). It should be noted that, as shown in the representative drawings below, children exhibit two completely different tendencies toward drawing this theme: one is to express it through graphics, and the other is to express it through mind maps (Fig. 18). However, from the perspective of semantic similarity, the difference between these two forms of expression does not affect the understanding of semantics by the word2vec algorithm, because even when using graphics to represent this theme, a considerable number of students have added prompts in the drawings. The detailed information will be elaborated in the "Discussion" section of this article.

## Recognition by ChatGPT and human

Due to the overall accuracy rate being only 46.7%, there is a possibility that such a low accuracy may be due to the low image recognition ability of the LLM we have chosen? For this purpose, we conducted two additional experiments: validation from human and ChatGPT. We recruited a total of 46 adult participants and identified 225 drawings (accounting for 15.8% of the original 1420 drawings). In addition, a total of 284 drawings were identified using ChatGPT-4 (20% of the original 1420 drawings). The selection of drawings adopts a pseudo-random method, i.e. first use the RANDBEATWEEN command in Excel to generate a column of pseudo-random numbers, which correspond to the serial numbers of children's drawings, and then we select the drawings for recognition and verification based on the pseudo-numbers.

Since we used a sampling method, we first tested whether a sampling rate of 15.8% and 20% could represent the overall results. We retrieved the accuracy data of ChatGLM for recognizing 225, 284, and all 1420 children's drawings and verified the difference between the two samplings and population using chi-square test. The result of "225" is $\chi^2(1) = 1.759$, $P = .185$. The result of "284" is $\chi^2(1) = 0.992$, $P = .319$. So, there are insufficient evidence of significant differences, indicating that a sampling rate of 15.8% and 20% can basically explain the overall situation.

The results of the accuracy comparison are shown in Table 2 and Fig. 19. It can be seen that the performance of humans and the two LLMs in children's drawing recognition accuracy is very close. To further validate this intuitive feeling, we conducted a one-way ANOVA analysis on the four columns of data in Table 2. The results showed that, $F(3, 32) = 0.217$, $P\ (3, 32) = .884$. There is almost no significant difference.

Therefore, overall, we believe that the choice of LLM does not affect the conclusion of this article.

## Semantic similarity between human and LLM

If the accuracy of human and LLM image recognition is similar, are their reasons for image recognition also very similar? Therefore, we recruited human participants to recognize more children's drawings, in order to have the opportunity to compare the diversity of language generated by humans and LLM. A total of 46 adult participants were recruited this time, and 225 children's drawings were identified. As LLM participants, ChatGPT-4.0 also recognized the same drawings.

Figure 20 compares the quartiles of semantic similarity between humans and GPT, while Fig. 21 compares the distributions. It can be clearly seen that the semantic similarity of human language is significantly lower than that of machines.

However, what could be the reason for this difference? We have made two speculations: one possibility is that human language generation ability is stronger than LLM, i.e. humans can generate and organize language using richer expressions. Another possibility is that human language contains richer meanings, and LLM may just be honestly " discuss it literally." Of course, it is also possible to have both.

To verify these two possibilities, we first conducted word frequency statistics on both human and GPT generated languages to verify whether their word frequencies conform to the Zipf distribution. The results are shown in Table 3 and Fig. 22 (left). It can be seen that there are indeed some differences between the two, but they are not obvious. The independent $t$-test confirmed this point ($F(16) = 0.001$, $P\ (16) = .223$). It can be seen that there is not much difference between human language and GPT.

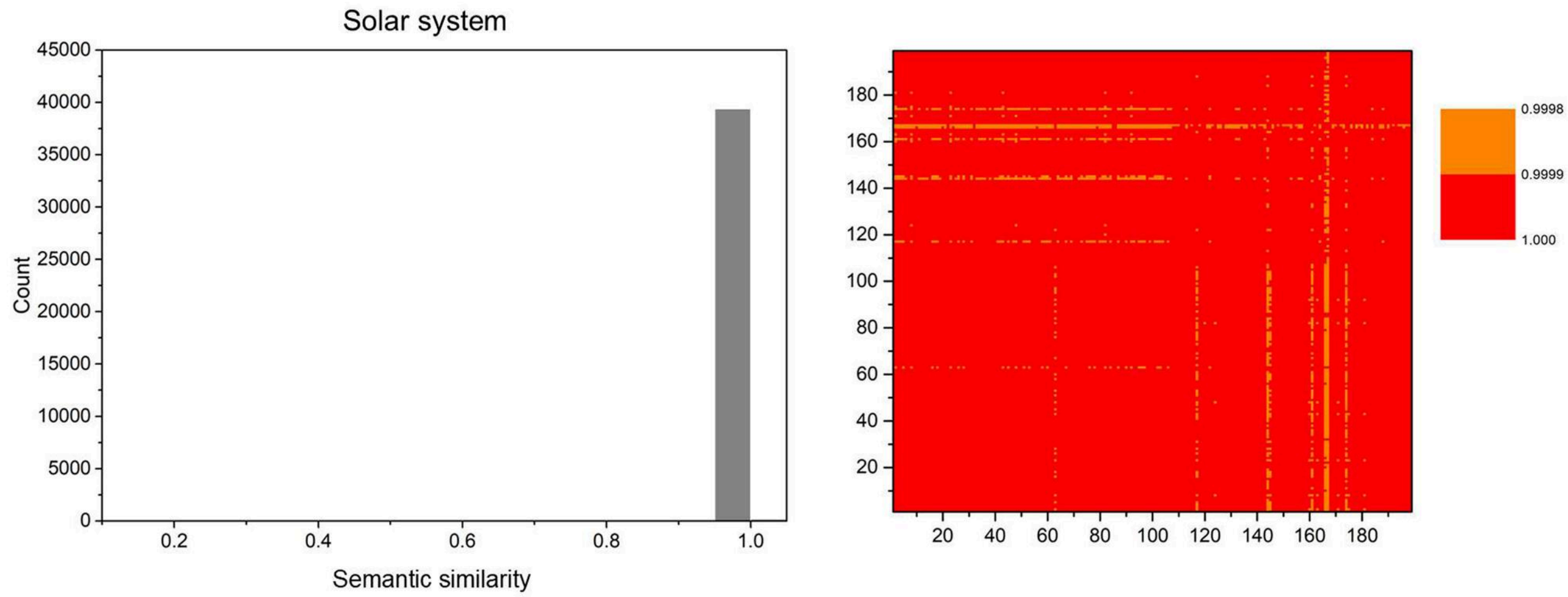

**Figure 7** Distribution diagram (left) and heatmap (right) of semantic similarity of the concept of "Solar system"

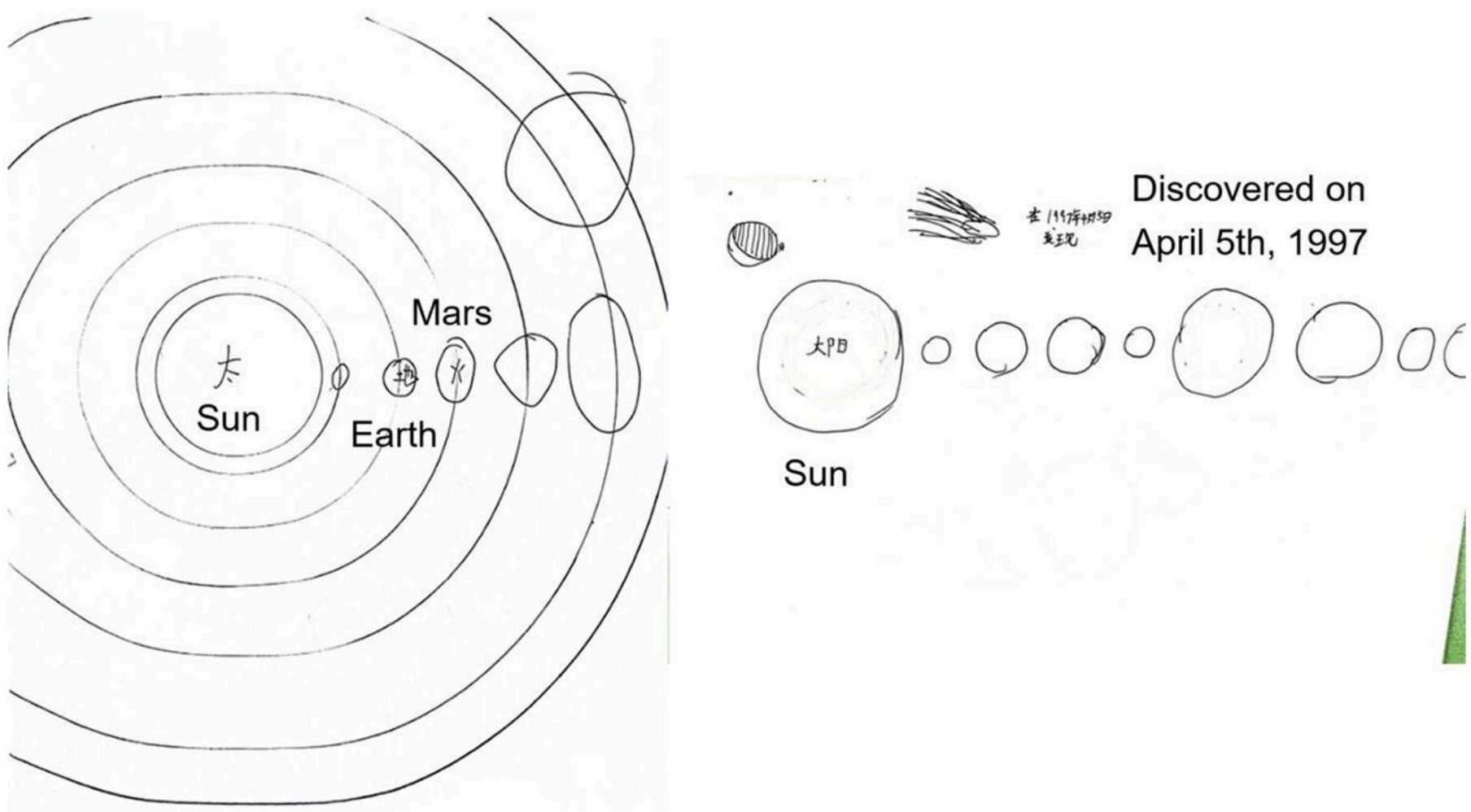

**Figure 8** Representative drawings of the concept of "Solar system"

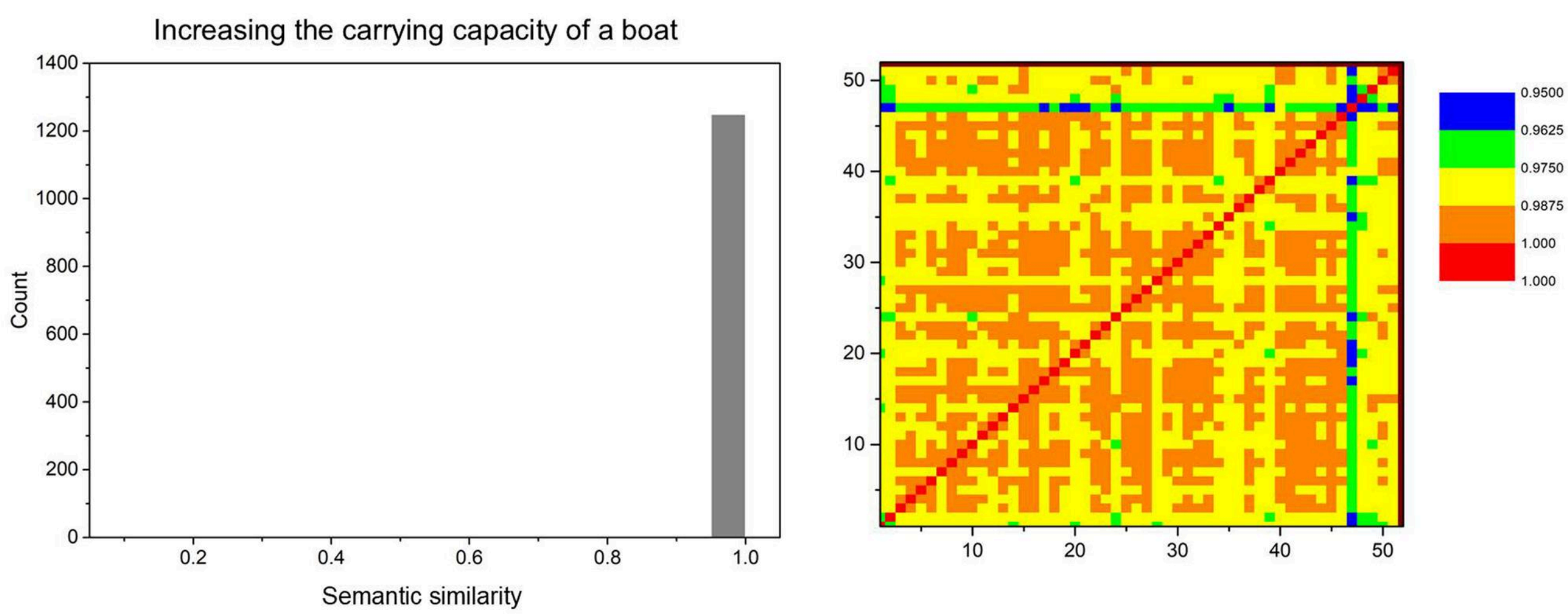

**Figure 9** Distribution diagram (left) and heatmap (right) of semantic similarity of the concept of "Increasing the carrying capacity of the boat"

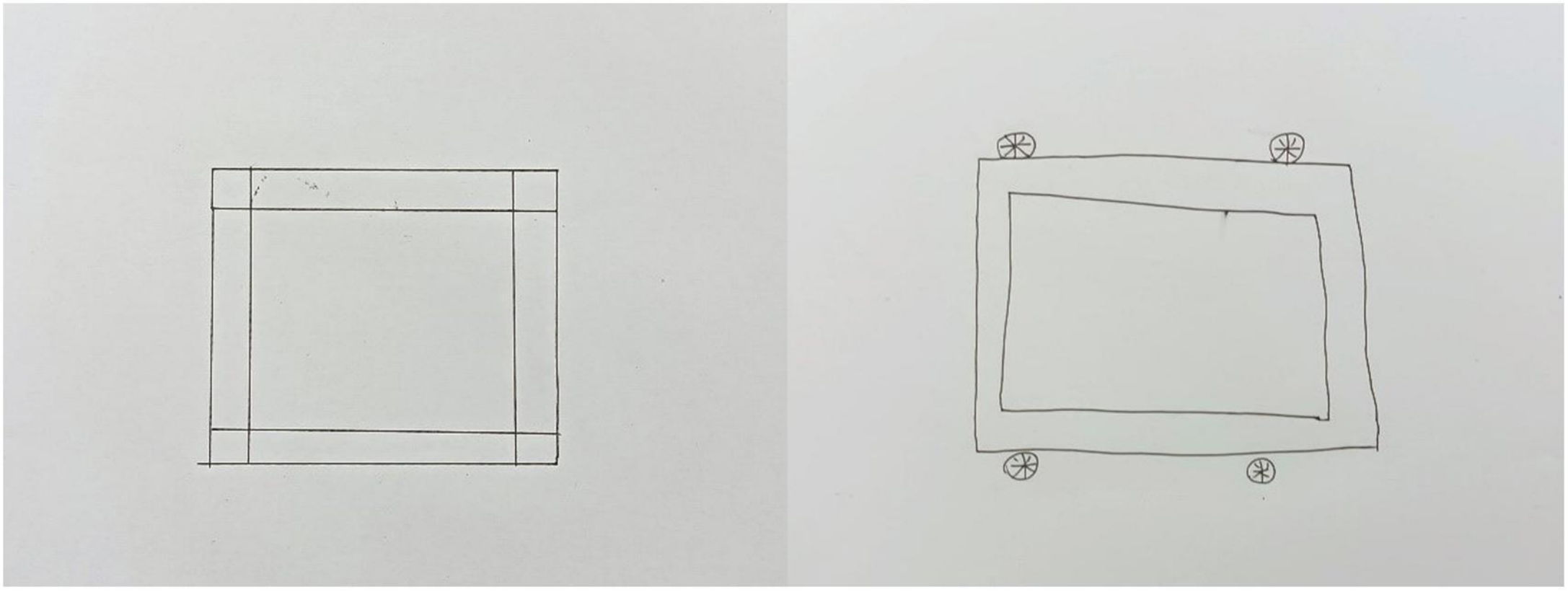

**Figure 10** Representative drawings of the concept of "Increasing the carrying capacity of a boat" (the four circular objects in the drawing on the right were small iron hoops used to increase the weight)

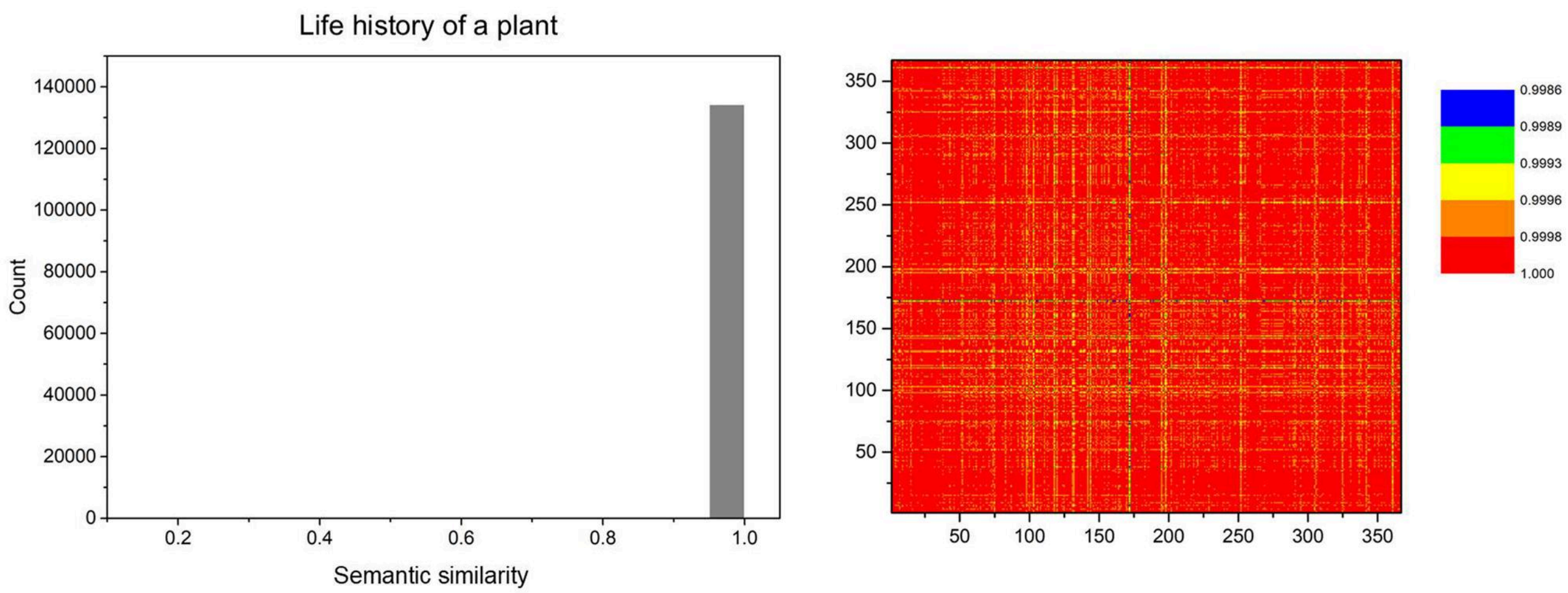

**Figure 11** Distribution diagram (left) and heatmap (right) of semantic similarity of the concept of "Life history of a plant"

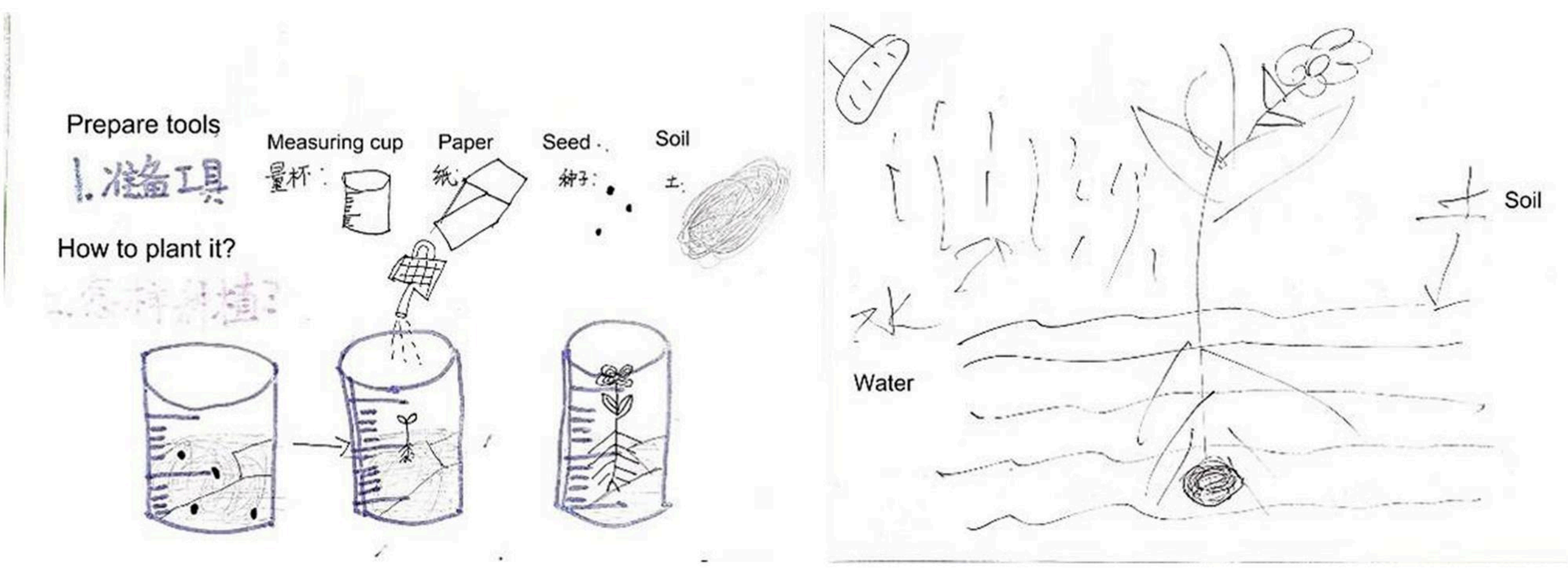

**Figure 12** Representative drawings of the concept of "Life history of a plant"

To verify the second hypothesis, we used LDA algorithm to explore the differences between human and GPT languages in terms of latent topics, as latent topics can well represent the diversity of text meanings.

To this end, we first used LDA algorithm to mine the potential themes of human and GPT generated text of each concept and listed the top 10 main themes. Then, using the method of information entropy, the entropy values of these topics were calculated to represent the diversity of topics in different texts. In order to ensure the stability of the data and comparability of the results, we fixed the parameters of LDA in filtering the text as follows: no_before = 3, no_before = 0.8, and keep_$n$ = 100 000. The comparison of information entropy between them is given in Table 4 and Fig. 22 (right).

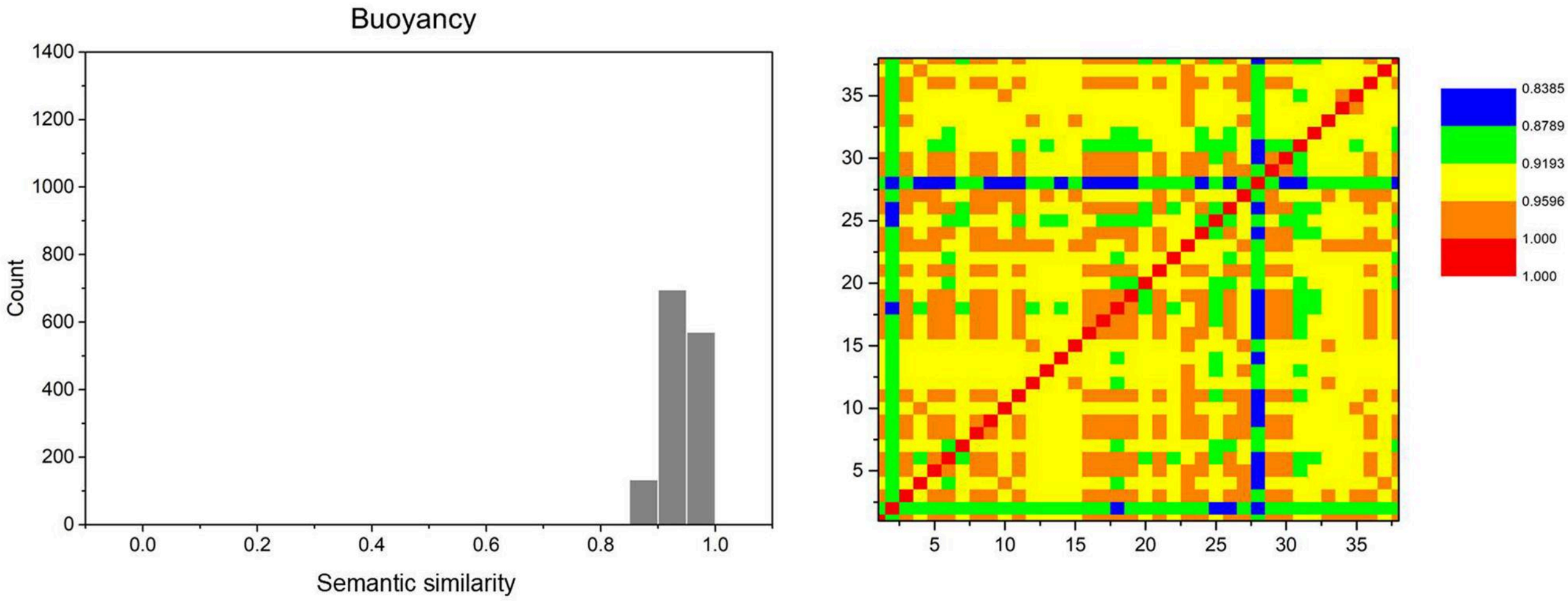

**Figure 13** Distribution diagram (left) and heatmap (right) of semantic similarity of the theme of "Buoyancy"

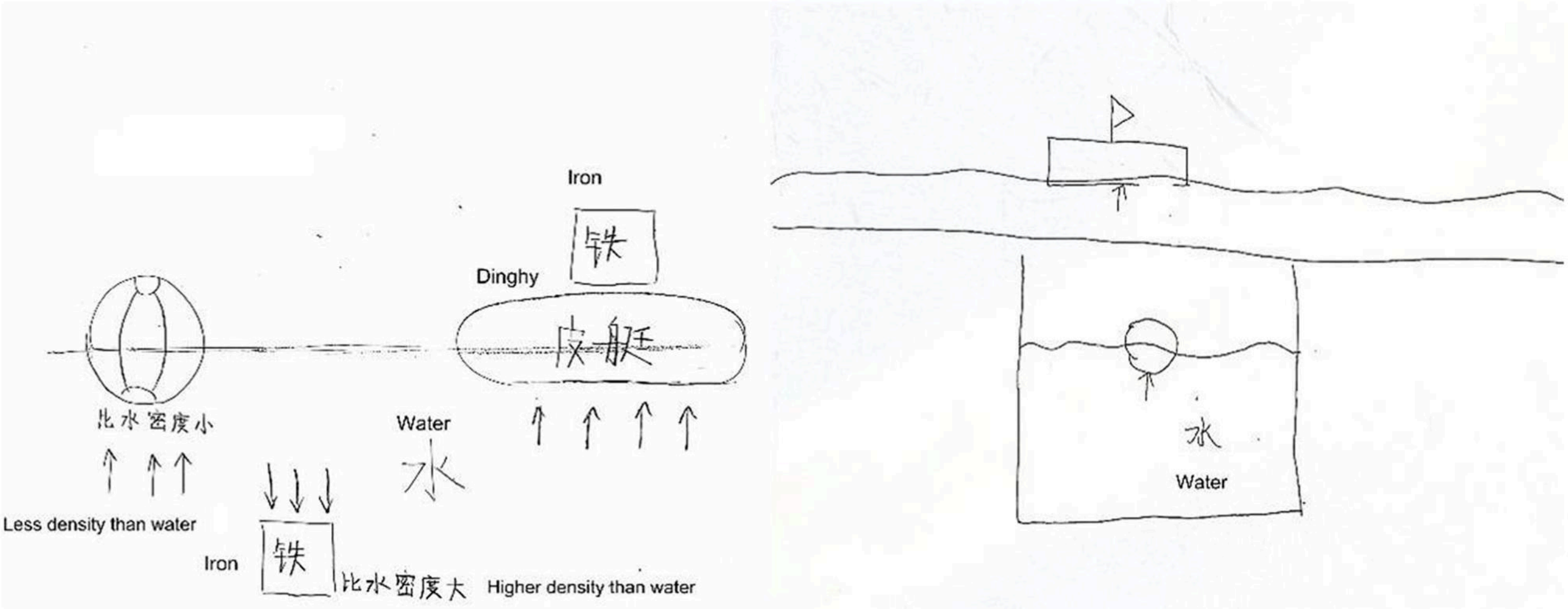

**Figure 14** Representative drawings of the theme of "Buoyancy"

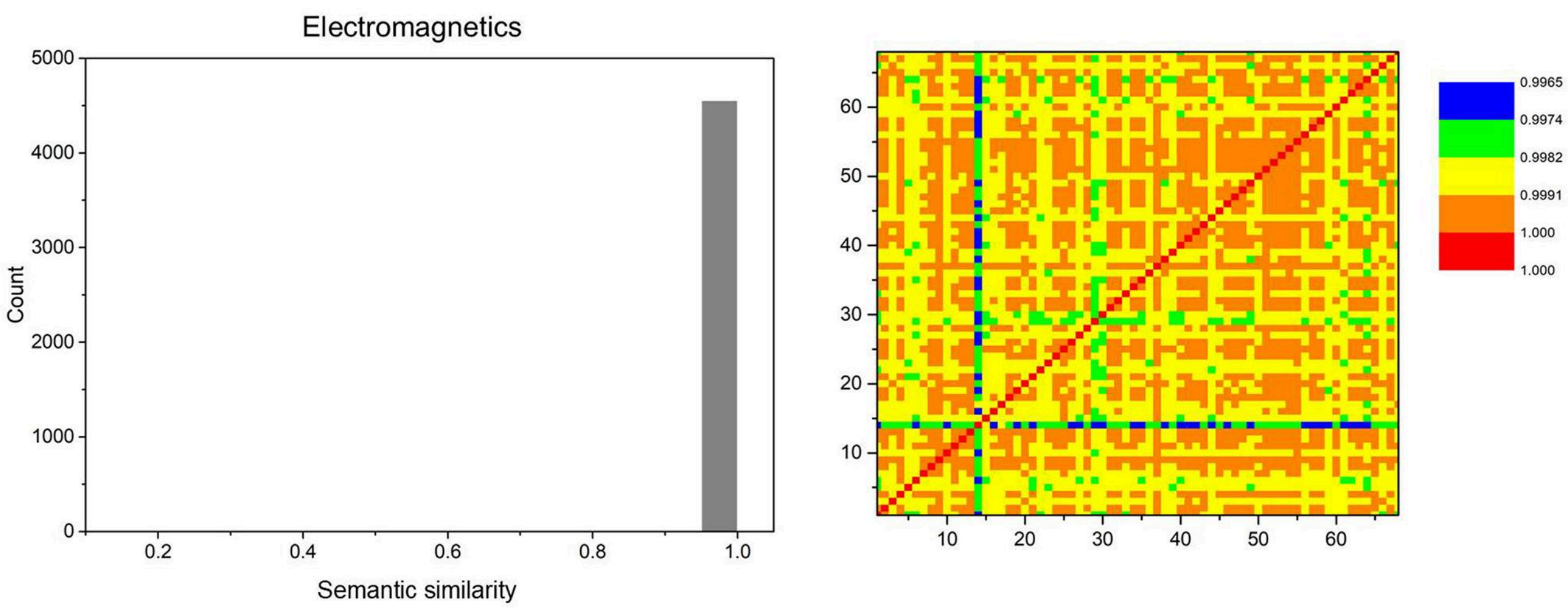

**Figure 15** Distribution diagram (left) and heatmap (right) of semantic similarity of the concept of "Electromagnetics"

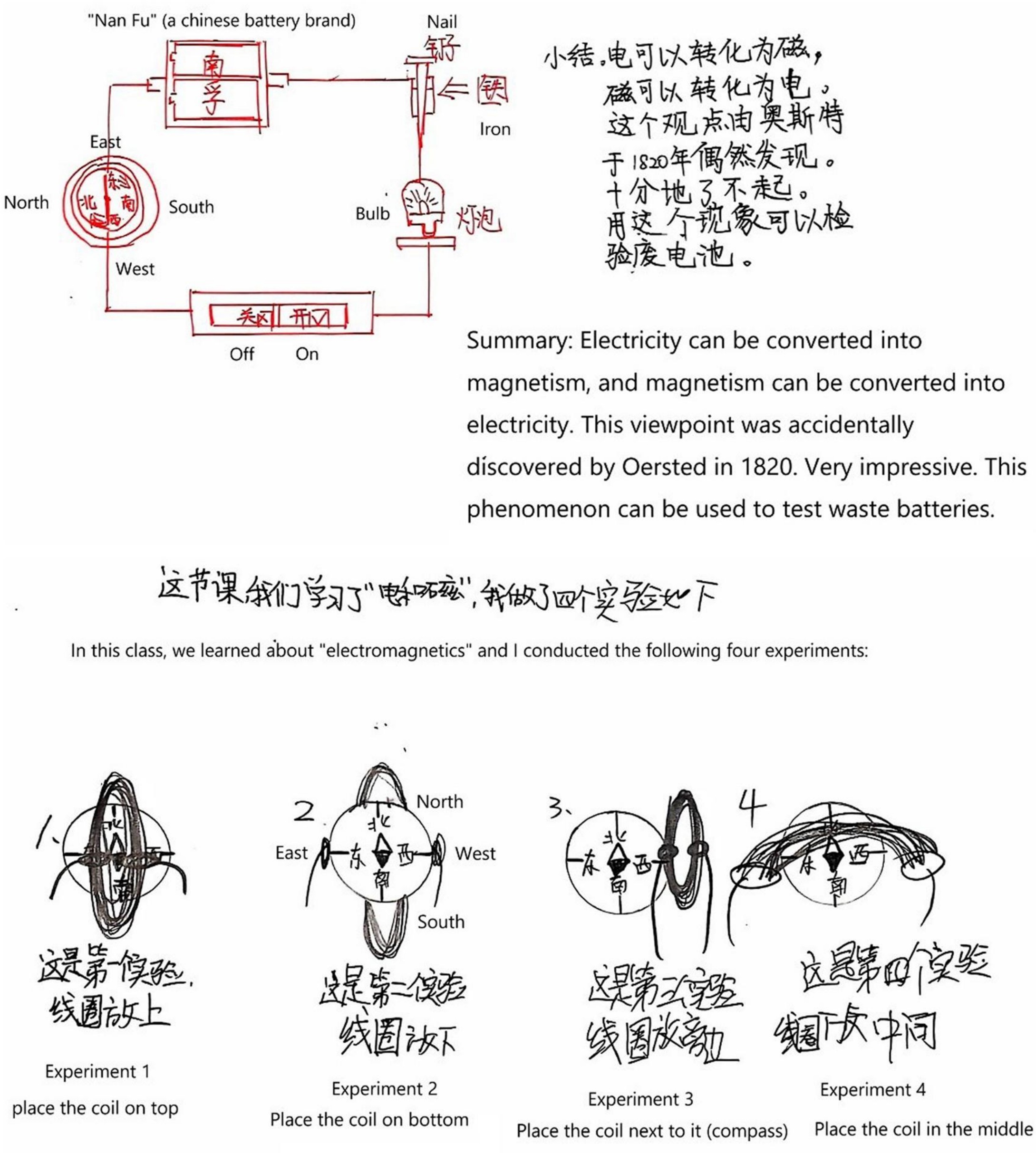

**Figure 16** Representative drawings of the concept of "Electromagnetics"

It can be seen that the entropy value of human language is significantly higher than GPT (average entropy $H_{GPT} = 1.459$, $H_{hum} = 2.188$, $t$-test; $P < .001$), confirming the second hypothesis that the diversity of potential topics in human language is significantly higher than that of machines.

By reviewing the reasons for human and machine drawing recognition, we found that there are at least three aspects that may confirm the above results: firstly, the diversity of human participants composition. About one-third of all 46 participants were science teachers, who not only had significant differences in their language compared to other participants, but also had a phenomenon of judging children's drawings. Such as "The diagram depicts the major celestial bodies in the solar system, indicating that students have a relatively rich understanding of the solar system," which is impossible to see in the generative language of LLM. The second is the semantic diversity caused by rich experiences. There is a certain degree of similarity between "Life History of a Plant" and another lesson "Seed Germination," so a considerable number of human participants confuse these two lessons, while LLM does not have confusion due to the lack of similar experiences. The third reason is other factors. Such as human general knowledge is richer than LLM (one human participant misunderstood a child's drawing as one of the Chinese mahjong tiles, but LLM may don't know what is the mahjong and naturally will not think in that direction), or homophone lead to misleading tokenizers and algorithms.

The above results further demonstrate the value of using machines for this study: the results of human participants are greatly influenced by individual heterogeneity, but machines almost do not have this problem. Although the results of machines may not be superior to those of humans, their results have strong data stability, are easier to reproduce, are more friendly to statistical analysis, and are more suitable for mining patterns from them.

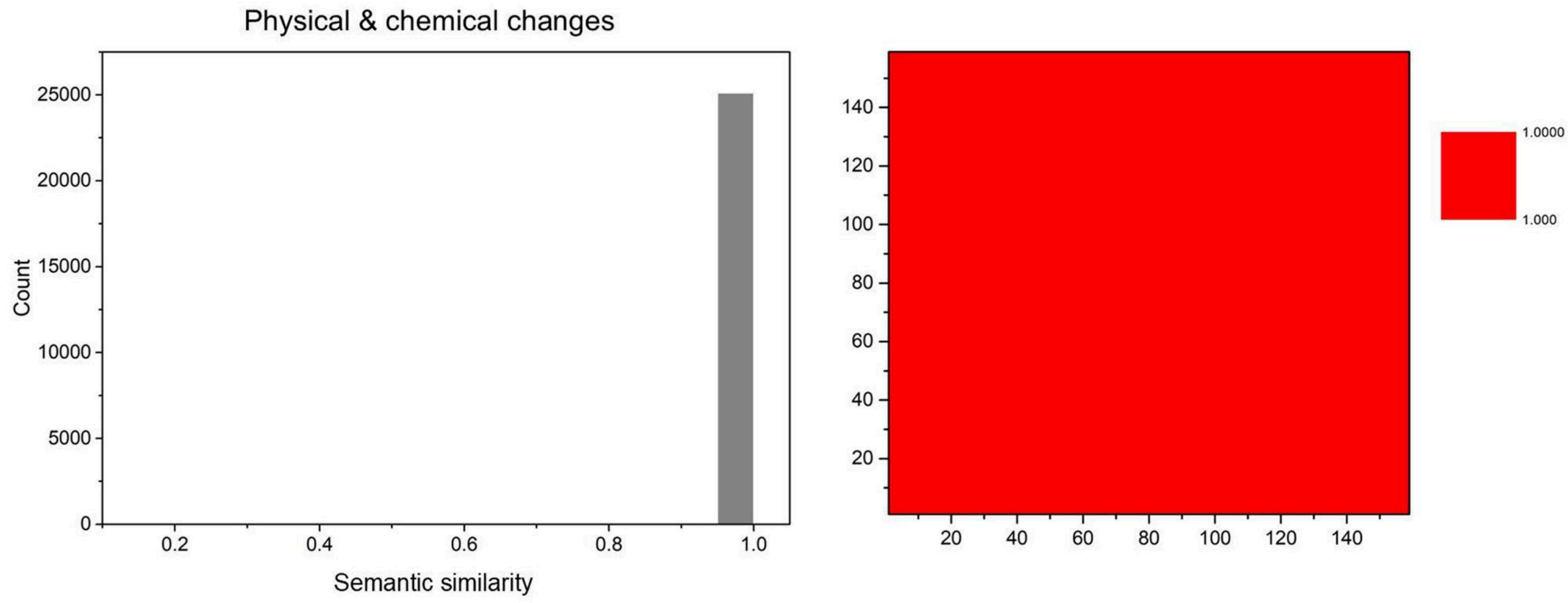

**Figure 17** Distribution diagram (left) and heatmap (right) of semantic similarity of the concept of "Physical and chemical changes"

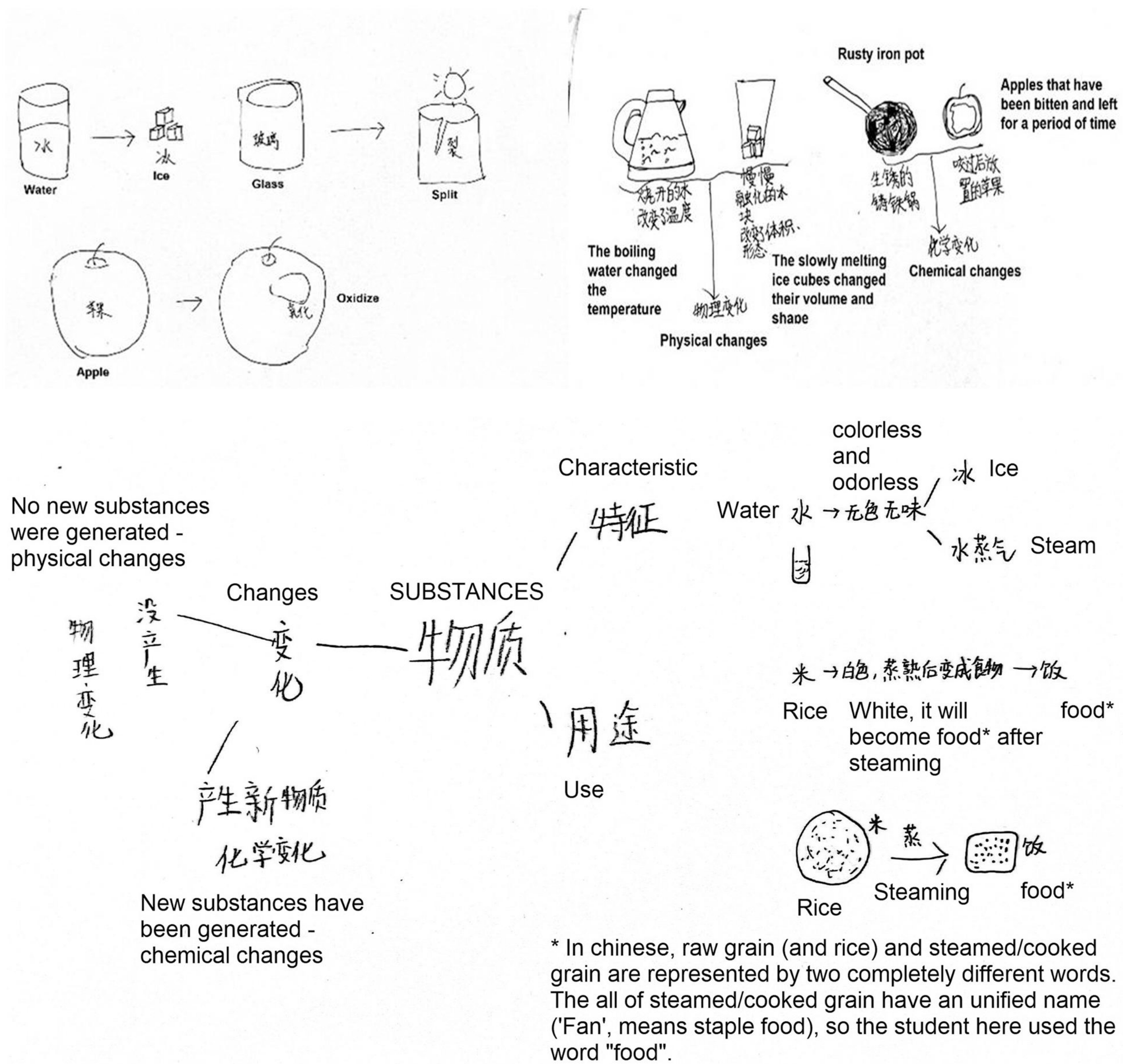

**Figure 18** Representative drawings of the theme of "Physical and chemical changes"

## Analysis of related factors

As mentioned above, this article uses Kendall rank correlation coefficient (Kendall's τ) to analyze the factors that affect drawing's representations.

### *Relationship with sample sizes*

Is there a relationship between the above results and the total sample size? We separately calculated the correlation coefficients between "sample size" and three factors ("accuracy," "median value of semantic similarity," and "75% Region"), and the results showed that the correlation coefficients of the three factors were 0.611, 0.609, and −0.493, respectively (the three bar charts above Fig. 24). It can be seen that sample size has a strong correlation with accuracy and median value of semantic similarity and has a moderate correlation with semantic similarity concentration, and this moderate correlation is not significant $P = .07$).

**Table 2.** Comparison of human, ChatGPT, and ChatGLM.

| | Human | ChatGPT | ChatGLM | ChatGLM-total |
|---|---|---|---|---|
| **Circuit** | 0.795652 | 0.839286 | 0.892857 | 0.763 |
| **Electromagnetics** | 0.967391 | 0.727273 | 0.818182 | 0.514 |
| **Boiling** | 0.217391 | 0 | 0 | 0 |
| **Buoyancy** | 0.608696 | 0.666667 | 0.5 | 0.5 |
| **Solar eclipses** | 0.747283 | 0.391304 | 0.552174 | 0.44 |
| **Increasing** | 0.086957 | 0 | 0 | 0.033 |
| **Solar system** | 0.9375 | 0.843137 | 0.45098 | 0.65 |
| **Physical and chemical** | 0.86087 | 0.694444 | 0.75 | 0.513 |
| **Life history** | 0.041667 | 0.212121 | 0.575758 | 0.712 |
| **Average** | 0.605978 | 0.559859 | 0.51555 | 0.467222 |

Among them, "ChatGLM total" represents the recognition accuracy of each concept in the original 1420 children's drawings by the LLM, while "ChatGLM" represents the recognition accuracy of the 284 children's drawings in the additional experiment corresponding to the LLM.

### *Relationship with the abstract degree of themes/concepts*

Based on J. Piaget's theory of developmental stages in children, the abstract thinking ability of elementary school students has not yet been fully developed [20]. Therefore, we speculate whether the above results are related to the degree of abstraction of concepts? That is to say, the higher the degree of abstraction, the lower the accuracy and the lower the semantic similarity? However, how to define the degree of abstraction of concepts? Children and adults have different understandings of abstraction. Some adults believe that concrete concepts (such as mathematical laws of force or motion) are abstract and difficult to grasp in the eyes of children because they cannot be directly seen [21].

Therefore, we have designed and implemented a new experiment here. We have recruited 10 new raters to complete a 5-point Likert scale (very abstract, abstract, neutral, concrete, very concrete). These participants all have over 7 years of teaching experience, and 4 of them hold the title of "Senior Teacher" of science education. The overall results are shown in Table 5, assign the 1 to 5 points from "very abstract" to "very concrete," respectively.

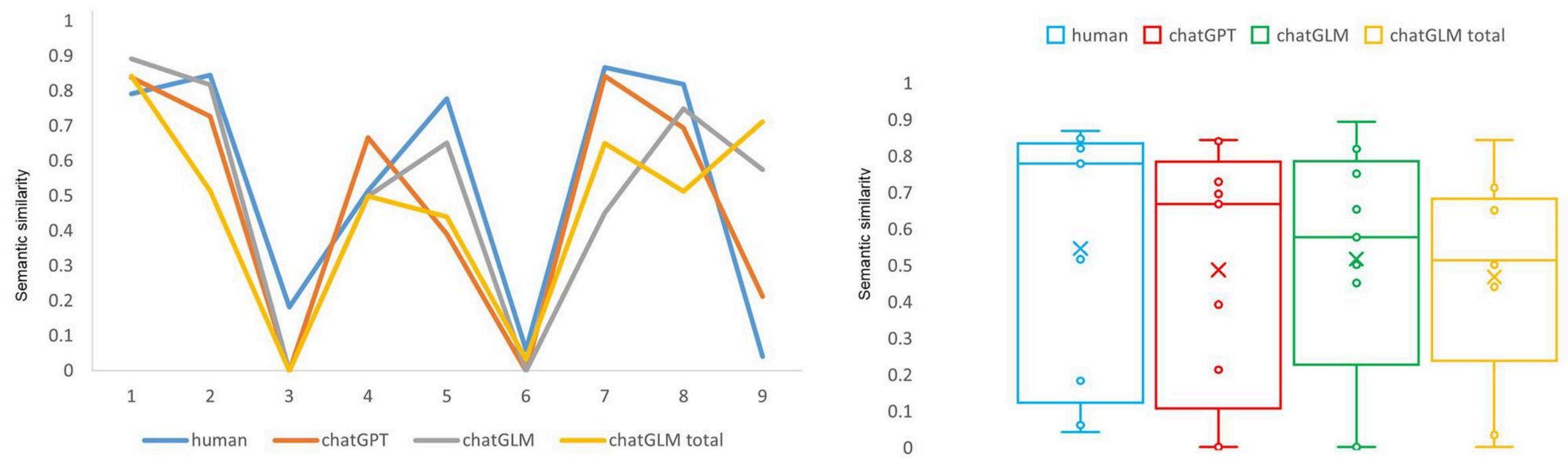

**Figure 19** The performance of humans and LLM in identifying nine concepts. The horizontal axis 1–9 in the left figure corresponds to nine concepts, respectively

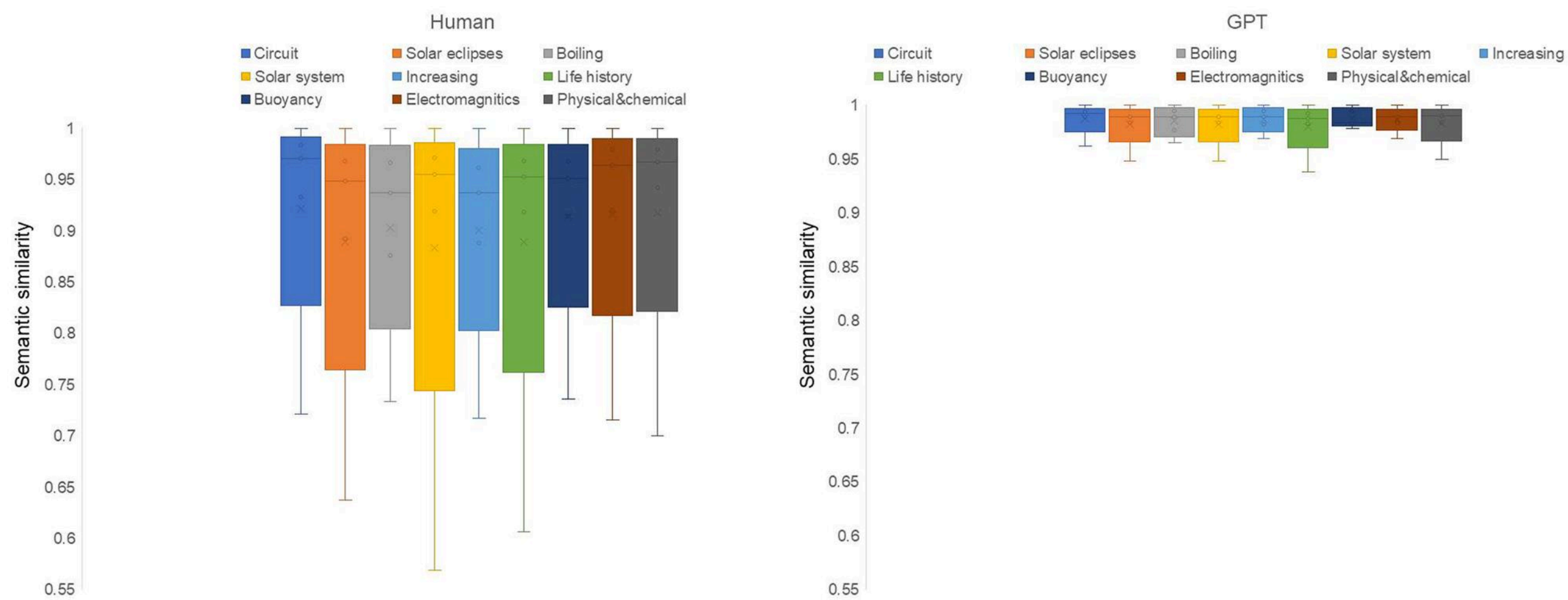

**Figure 20** The quartiles of semantic similarity between humans and GPT

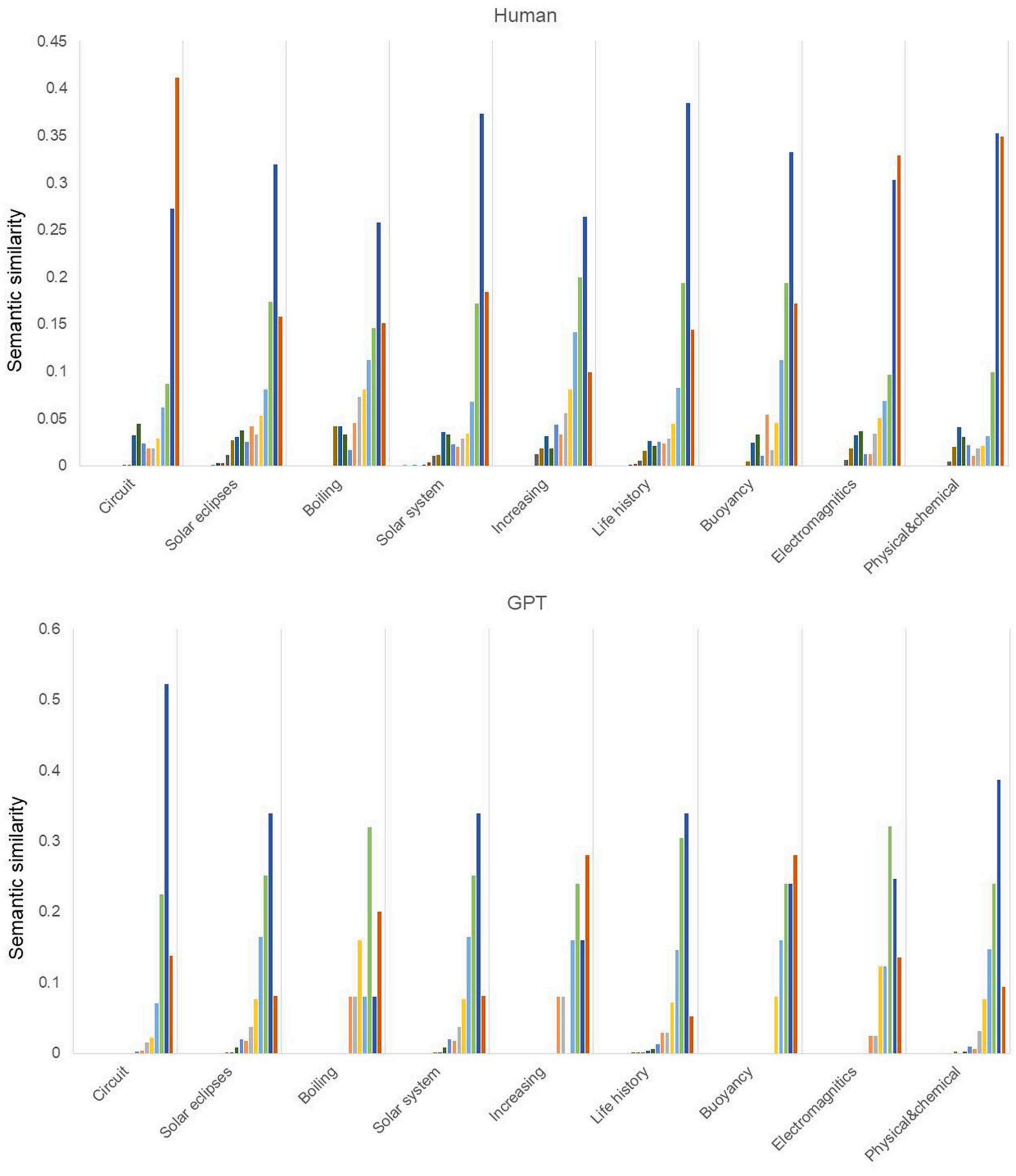

**Figure 21** The semantic similarity distributions between humans and GPT

**Table 3.** The Zipf values between humans and GPT.

| | GPT | Human |
|---|---|---|
| **Circuit** | 1.168 | 1.337 |
| **Solar eclipses** | 1.145 | 1.292 |
| **Boiling** | 0.75 | 0.856 |
| **Solar system** | 1.152 | 1.304 |
| **Increasing** | 0.738 | 0.733 |
| **Life history** | 1.061 | 1.101 |
| **Buoyancy** | 0.829 | 1.021 |
| **Electromagnetics** | 0.859 | 1.022 |
| **Physical and chemical** | 1.055 | 1.131 |

Figure 23 is a bar chart of the average ratings of nine concepts, with the red horizontal line representing the average level of ratings for all concepts.

We used the within the Intraclass Correlation Coefficient (ICC) instead of Cohen's Kappa test for consistency testing. The results showed that the Cronbach's alpha $\alpha = 0.618$, ICC(3,1) = 0.139, ICC

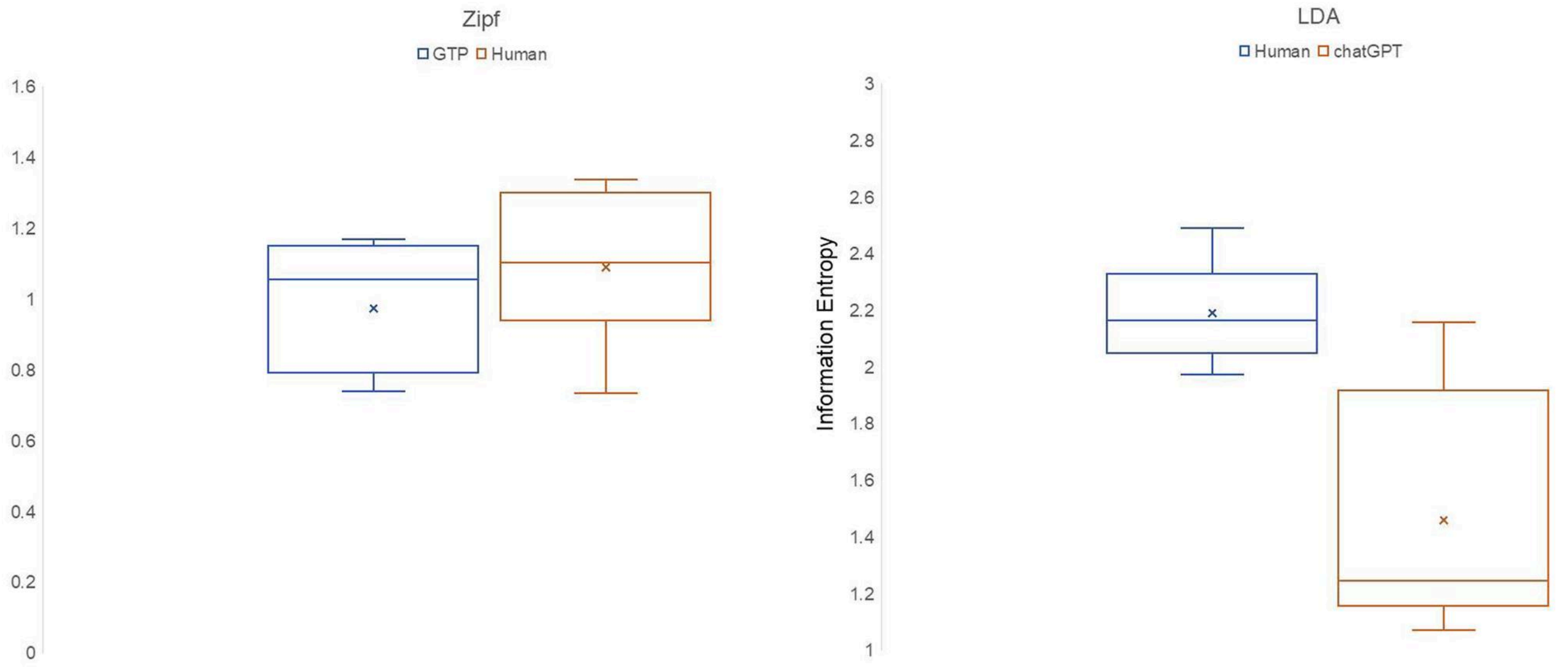

**Figure 22** The results about Zipf (left) and LDA (right) between humans and GPT

**Table 4.** The information entropy of LDA between humans and GPT.

| | GPT | Human |
|---|---|---|
| **Circuit** | 2.163 | 1.243 |
| **Solar eclipses** | 2.108 | 1.238 |
| **Boiling** | 2.489 | 1.867 |
| **Solar system** | 2.058 | 1.07 |
| **Increasing** | 2.378 | 1.965 |
| **Life history** | 2.035 | 1.205 |
| **Buoyancy** | 2.279 | 2.156 |
| **Electromagnetics** | 1.973 | 1.283 |
| **Physical and chemical** | 2.208 | 1.102 |

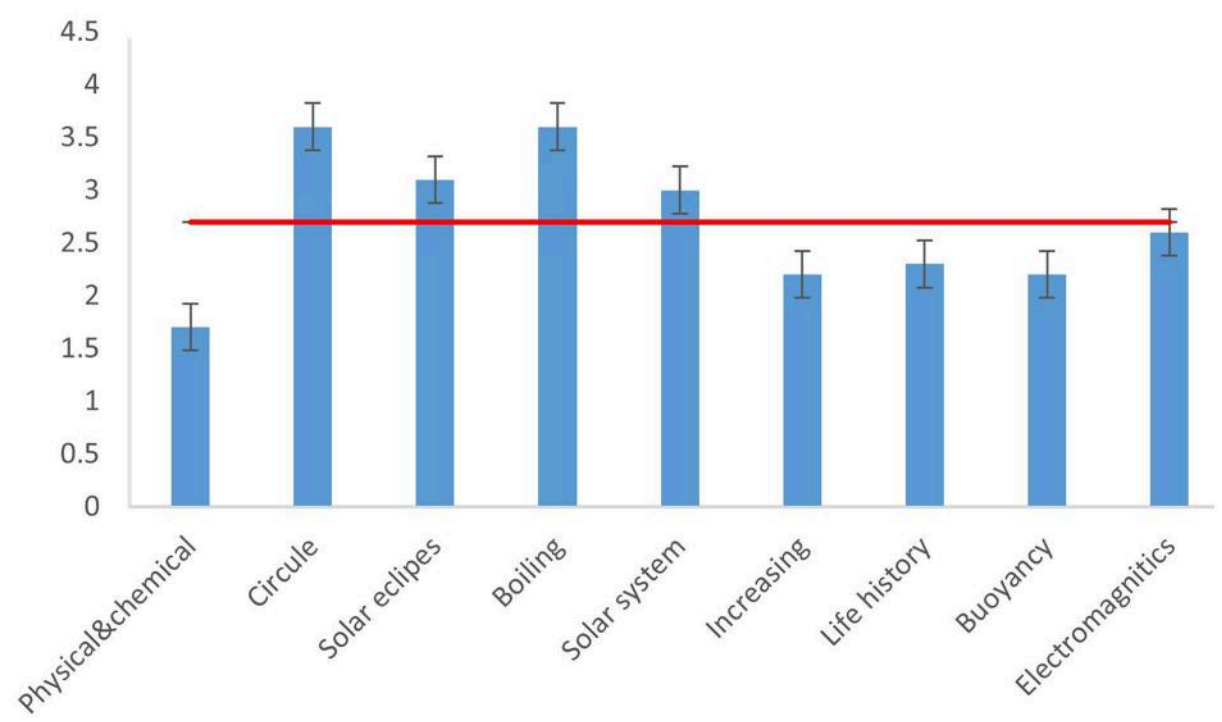

**Figure 23** The result of scores about each concept

**Table 5.** The result of five-point Likert scale.

| | Raters | | | | | | | | | |
|---|---|---|---|---|---|---|---|---|---|---|
| | 1 | 2 | 3 | 4 | 5 | 6 | 7 | 8 | 9 | 10 |
| **Physical and chemical** | 4 | 1 | 3 | 1 | 1 | 1 | 1 | 1 | 1 | 3 |
| **Circuit** | 5 | 3 | 2 | 4 | 4 | 4 | 4 | 5 | 1 | 4 |
| **Solar eclipses** | 2 | 4 | 2 | 2 | 5 | 5 | 1 | 5 | 1 | 4 |
| **Boiling** | 5 | 4 | 4 | 1 | 5 | 5 | 1 | 4 | 2 | 5 |
| **Solar system** | 3 | 4 | 3 | 2 | 4 | 4 | 1 | 4 | 2 | 3 |
| **Increasing** | 3 | 2 | 4 | 3 | 1 | 1 | 3 | 1 | 2 | 2 |
| **Life history** | 3 | 2 | 2 | 3 | 1 | 1 | 1 | 2 | 4 | 4 |
| **Buoyancy** | 1 | 4 | 2 | 3 | 1 | 4 | 1 | 1 | 1 | 4 |
| **Electromagnetics** | 3 | 1 | 4 | 2 | 5 | 2 | 5 | 1 | 1 | 2 |

(3,9) = 0.618. The overall behavior of the 10 raters shows a moderate to strong consistency.

According to the results in Table 5, we calculated Kendall's $\tau$ between the Likert mean score of each concept and its "75%," median, and accuracy. The results were −0.03 ($P = .914$), −0.029 ($P = .916$), and 0.057 ($P = .833$), respectively. This result suggests that the level of abstraction of concepts may not be significantly correlated with all these factors.

### *Reproducibility the contents in the classroom*

Another factor we are concerned about is whether students will express abstract themes/concepts by reproducing contents that have appeared in class through drawings? Due to the elusiveness of these themes/concepts for children, textbooks often present them through representative examples, such as express the "Life history of a plant" through Impatiens. So, will students represent these difficult things by reproducing the examples that have appeared in the classroom? We used the examples that appeared in the textbook as the basis and calculated the proportion of these example words in the LLM's recognition reasons through word frequency statistics and calculated the total proportion using weighted average (weighted by sample size). The results are as follows (Table 6):

The weighted average reproduction rate is 76.19%, which means that at least three-quarters of the students have reproduced the classroom content in the drawings more or less to represent these difficult things. Considering that LLM's recognition accuracy is not 100%, this proportion is likely to be even higher.

### *Relationship with the focus points*

This is what we discovered after carefully examining the drawings: we found that the concept of "boiling" originally required students to understand the process of water to steam change, but most of the students drew the experimental activities (i.e. the experimental apparatus). This experiment is an indispensable process for heating water, but not the purpose of the lesson. However, it was used by students as a representation of the phenomenon of water boiling. In other words, there is a significant

**Table 6.** Reproduction rate of samples encountered in the classroom.

| Themes/concepts | Sample size | Count of reproduction | Rate of reproduction (%) | Examples[b] |
|---|---|---|---|---|
| **Increasing the carrying capacity of a boat** | 52 | 24 | 46.15 | Square, rectangle |
| **Life history of a plant** | 368 | ——[a] | —— | —— |
| **Buoyancy** | 40 | 8 | 20 | Bottle, submarine |
| **Physical and chemical changes** | 70 | 160 | 100 | Water, salt, flour, dough, iron pot, rust, eggs, apples, ice cubes |

[a] It is truly difficult for us to identify whether the plants in drawings were Impatiens or not (most of the drawings do not provide the names of the plants), so we did not calculate the data for this theme.
[b] Due to the large number of examples appearing in this lesson, it is not required that all examples be drawn. As long as one example appears in the drawings, it will be counted.

**Table 7.** Coding and reasons for nine concepts about the relationship between experimental results and teaching purpose.

| Themes/concepts | Codes | Reasons of codes |
|---|---|---|
| **Circuit** | 1 | The experimental results are consistent with the teaching purpose, both pointing to the shape of the circuit (a "circle") |
| **Solar eclipses** | 1 | There is no experiment in this class |
| **Boiling** | 2 | The experiment is a necessary condition for boiling, but it is not the teaching purpose (i.e. it is not the boiling phenomena and changes in water) |
| **Solar system** | 1 | There is no experiment in this class |
| **Increasing the carrying capacity of a boat** | 2 | This lesson contains two activities, the first of which is not related to the teaching purpose and only about creating a small boat that can boat in the water |
| **Life history of a plant** | 1 | There is no experiment, just a summary of previous plant planting activities |
| **Buoyancy** | 1 | The experimental results are consistent with the teaching purpose, both pointing to regulating the sinking and floating of the small boat |
| **Electromagnetics** | 1 | The experimental results are consistent with the teaching purpose, both pointing to electromagnetic phenomena |
| **Physical and chemical changes** | 1 | The experimental results are consistent with the teaching purpose, both pointing to demonstrating a physical or chemical change |

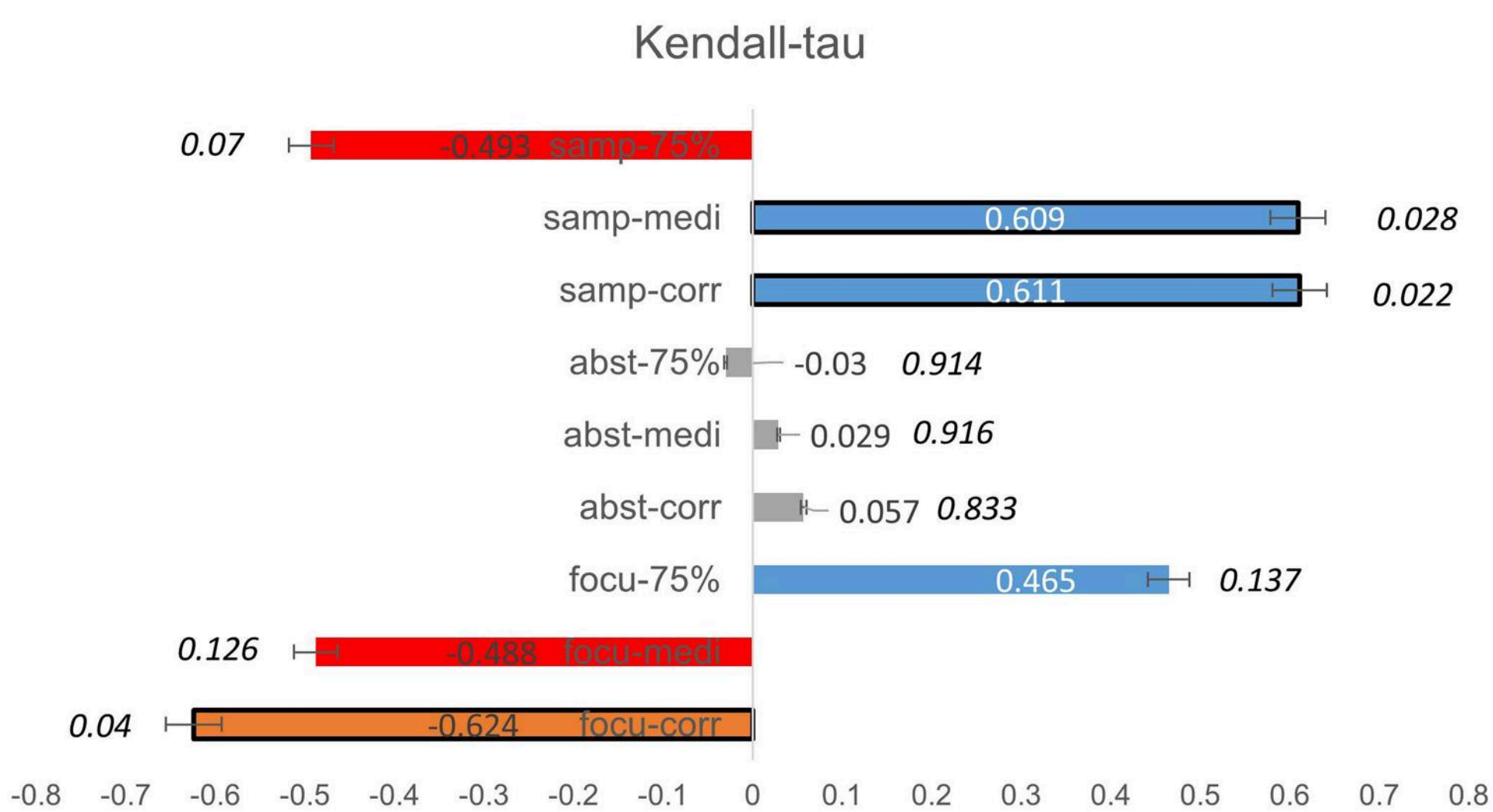

**Figure 24** The Kendall rank correlation coefficient results for each factor, where "samp" represents the factor of sample size, "abst" represents the factor of abstract degrees, "focu" represents the factor of focus points, "75%" represents 75% Region, "medi" represents median, and "corr" represents accuracy; "samp-medi" represents the correlation coefficient between the sample size and the median value (other similar); Color represents moderate or above correlation (tau > 0.4), and bold black boxes represent the correlation is significant (*P* < .05)., where italicized numbers represents the *P*-value of the correlation coefficient

deviation between the students' focus and the purpose in the textbook.

Therefore, we developed a coding rule based on the role of experiments in lessons and used it as an independent variable to conduct correlation analysis on the above three factors. The coding rules are as follows: those themes without experiments or experimental results that were equivalent to the teaching purposes will be coded 1. The experimental results were not the teaching purposes, but only the indispensable process of teaching purposes will be coded 2 (simple drawing and hands activities are not considered experiments). The coding result is as follows (Table 7):

Due to the fact that the above coding rule involves a deep understanding of textbook content, we invited two senior science teachers (both of whom have senior professional titles) to review the above rules and received their unanimous agreement.

The correlation coefficients are −0.624, −0.488, and 0.465 (the three bar charts below Fig. 24), indicating that the consistency between experimental results and teaching purposes can explain the accuracy very well. Specifically, it can effectively solve the problem of "collective bias" in students' drawings: students are more concerned with the experiment itself, but if the experiment deviates from the teaching purposes, students are likely to choose to remember the experiment rather than the real purposes. Meanwhile, there is a moderate correlation with the other two items (median value of semantic similarity value and 75% Region).

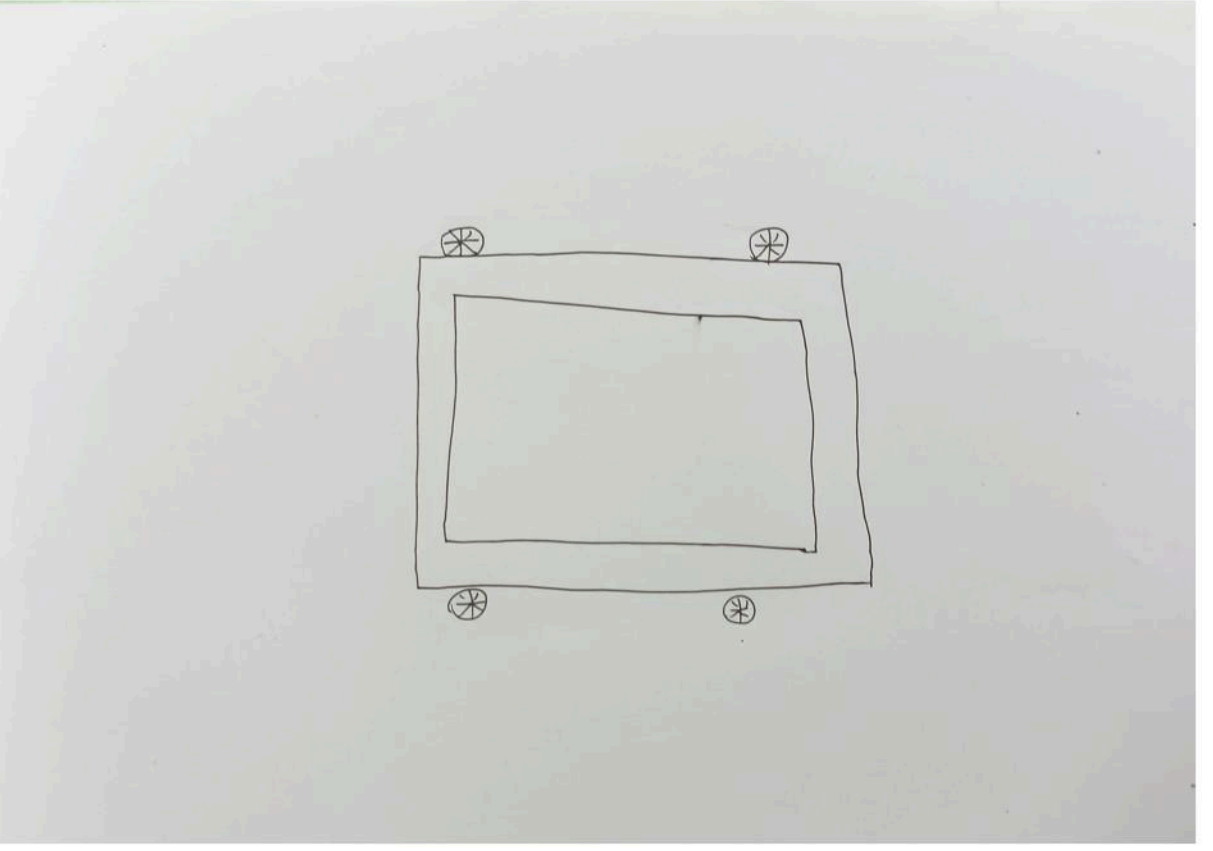

**Figure 25** It's hard to imagine that the student is drawing a small boat in this picture

## Discussion

This article outlines the representations of nine scientific concepts in children's eyes through 1420 children's scientific drawings and explores whether there is a consistent representation in children's scientific drawings through LLM image recognition and semantic similarity analysis. The results show that the representation of most drawings has consistency, manifested as most semantic similarity > 0.8. At the same time, it was found that the consistency of the representation is independent of the accuracy (of LLM's recognition), indicating the existence of consistency bias. In the subsequent exploration of influencing factors, we used Kendall rank correlation coefficient to investigate the effects of "sample size," "abstract degree," and "focus points" on drawings and used word frequency statistics to explore whether children represented abstract themes/concepts by reproducing what was taught in class. It was found that accuracy (of LLM's recognition) is the most sensitive indicator, and data such as sample size and semantic similarity are related to it. The consistency between classroom experiments and teaching purpose is also an important factor, and many students focus more on the experiments themselves rather than what they explain. In addition, most children tend to use examples they have seen in class to represent more abstract themes/concepts, indicating that they may need concrete examples to understand abstract things. In summary, this article provides a novel approach to understanding children's drawings, which utilizes LLM to convert drawing information into more accurate linguistic information, allowing for in-depth research on children's conceptual learning using linguistic tools.

### About the low accuracy of LLM image recognition

The first thing that catches attention in the above results is that LLM's image recognition accuracy is not high (46.7%). The additional tests on humans and ChatGPT mentioned earlier have shown that this result is almost unrelated to the selection of LLM. So, how can this phenomenon be explained? We believe that in addition to the factor of children's low drawing ability (which is actually a very important factor), there is another factor that cannot be ignored: many children's drawings that demonstrate conceptual understanding shown the thinking process of children, rather than the results or products of thinking.

As shown in Fig. 25, a square box is drawn in the middle of the image, with four circular objects surrounding the box. At first glance, it is difficult for anyone to associate this drawing with a small boat. The background of this drawing is derived from the lesson "Increasing the carrying capacity of a boat." According to the textbook design, the students' activities in this lesson mainly include two items: first, folding a piece of aluminum foil into a square box without a lid, which can be used as an abstract version of a small boat. The second item is to use some circular iron rings (i.e. the four circular objects in the picture) to adjust the carrying capacity of the ship, ensuring that the small boat will not sink. The main intention of this lesson is to help students discover the relationship between the bottom area and height of a ship by adjusting the number of iron rings (because the total area of aluminum foil has already been fixed, changing the bottom area will inevitably change the height of the ship). However, it can be clearly seen from the picture that the students' thinking seems to have completely paused in the "making a small boat" item, with only the four circles leaving a little trace of "adjusting the load capacity." That is to say, this drawing is an unfinished product of thought, reflecting the process of students' learning and thinking.

However, LLM will definitely recognize the drawing as a mature and complete thinking result by default. And unfortunately, the images are the only source of information for LLM to recognize the drawings. And this deviation will naturally lead to a lower accuracy rate in image recognition. Moreover, from the above results, it can be seen that the accuracy of human image recognition is not much higher than that of LLM, indicating that human participants also assume that a drawing is a complete thinking product when they receive it.

### The contradiction between accuracy and consistency of representation

However, looking back at the entire result, we found a rather strange thing: some themes/concepts (solar eclipse, buoyancy, electromagnetics, and physical and chemical changes) have an accuracy rate close to 50%, but their semantics are very close to each other (median > 0.99). In this study, semantic similarity is used to represent the consistency of students' drawing representations. A high accuracy rate and high semantic similarity indicate that students are able to correctly and consistently represent the concept with drawings. A low accuracy rate but high semantic similarity indicates that students may have a problem of "consistency bias." So, how can we explain why the accuracy is close to 50% but the semantic similarity is high? The accuracy rate approaching 50% indicates that there should be significant differences in students' drawings (i.e. the number of correct and incorrect drawings is almost equal), why is the semantic similarity still high? This seems somewhat abnormal.

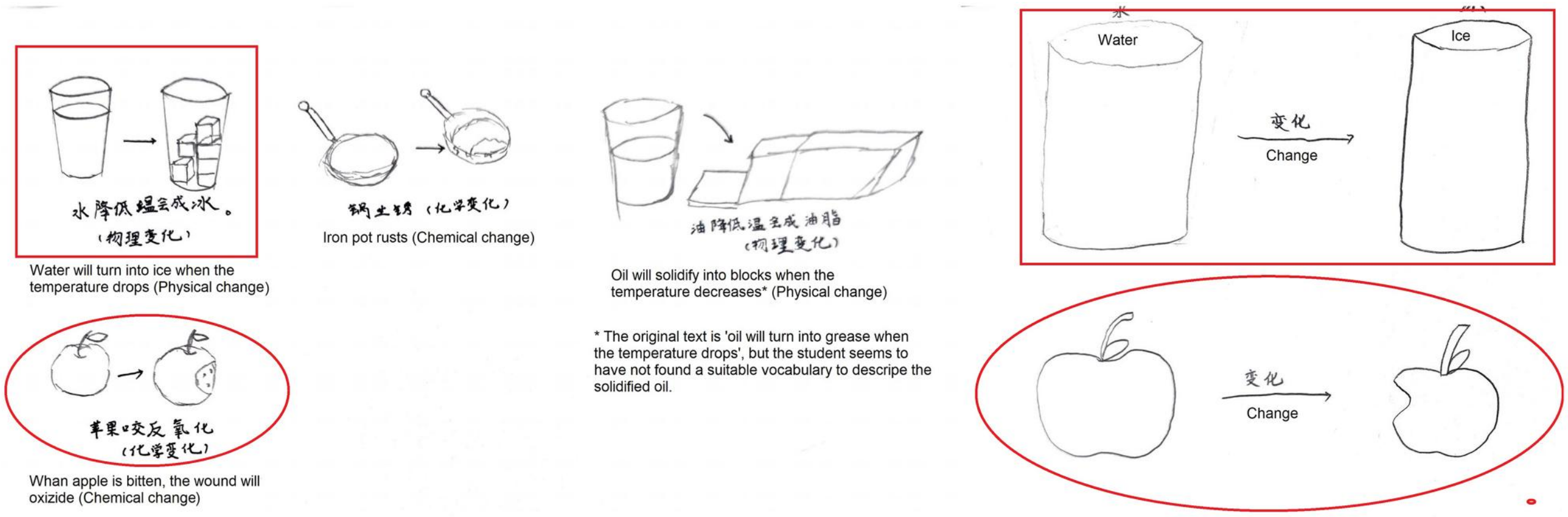

**Figure 26** Both left and right images depict the process of water turning into ice and the oxidation of apples

We carefully reviewed these drawings and found a possible "suspect": some drawings that LLM deemed incorrect actually used representations that were very similar to those of the correct drawings, but lacked some key prompts, thus misleading LLM. As shown in Fig. 26, both the left and right images depict the transformation of water into ice and the oxidation of apples after being bitten, but the difference is that the left image clearly indicates "physical changes" or "chemical changes" below each change, while the right image does not indicate these words. So, the left image was correctly recognized by LLM, but the right image was not.

## About abstract concepts

Another issue we are concerned about is how children understand abstract concepts, because obviously, abstract things are more difficult for people to learn concepts than concrete concepts, especially for children. For example, Elika Bergelson *et al.* [22] found that infants and young children have a significantly later understanding of abstract vocabulary than concrete vocabulary, but learning abstract things is an important task in education.

Previous research has focused on the following directions: firstly, utilizing language itself. For example, Gabriella Vigliocco *et al.* [23] proposed after reviewing previous research that children can use context to assist in understanding more abstract concepts. In fact, this is also the basic principle based on which mainstream NLP algorithm models are currently based [24]. In addition, Yanchao Bi *et al.* [25] found based on fMRI that semantic processing is also more involved in the cognition of abstract concepts of human. Secondly, using valence to understand [23], adults also use this effect to understand abstract concepts [26]. The third factor is the involvement of visual processing, as Amedeo D'Angiuli *et al.* [27] found that as abstract words are deeply processed, occipital lobe activity is significantly activated, indicating that children are likely to begin "imagining" the word.

However, for drawings, it is not entirely the same, as this approach clearly requires children to react and express their understanding of abstract concepts (i.e. they need to demonstrate it through behaviors), which is significantly different from the results obtained by directly scanning the brain using neuroscience equipment mentioned earlier. The previous review showed that there are two tendencies in children's drawing toward abstract concepts: one is influenced by language and the other is to search for prototypes that can represent this abstract concept (which seems to be similar to the visual processing mentioned earlier, as finding prototypes is a concrete process that clearly requires imagination [27]) [21].

We also found these two clear tendencies in this study, and we found that among the nine themes/concepts we explored, one theme clearly belongs to the abstract category: Physical and chemical changes (obviously, these two terms represent two major types of changes, rather than two concrete concepts). Therefore, the following analysis is mainly focused on this theme.

Firstly, as mentioned earlier, regarding the drawing of this theme, children have more or less reproduced the experiments in the classroom 100%. It can be seen that they have signs of searching for prototypes, because obviously the experiments once appeared in the classroom, which can naturally become typical exemplars for establishing prototypes of "Physical and chemical changes."

Secondly, the tendency to use language to represent this abstract theme is also very evident. Data shows that in 160 children's drawings, 53.8% of them contain at least one of prompts of "physical changes" and "chemical changes," and there are also many drawings that contain both. In addition, 89.5% of these drawings intentionally divide physical or chemical changes into two groups (left in Fig. 27). More representative is that 13.1% of the drawings do not have any graphics at all, but instead fully utilize mind maps to present this theme (right in Fig. 27). This seems to reflect that when students find it difficult to represent a concept with concrete graphics, they tend to use textual explanations as a concise and intuitive way to solve this problem. The data for judging the drawings mentioned above comes from two researchers who independently assessed three indicators of the drawings: Whether the drawing contains prompts for "physical changes" or "chemical changes"? Is the drawing intentionally grouping examples of the two changes? Is this drawing a graphic or a mind map? The Kappa coefficient is equal to 0.93 (95% CI [0.88, 0.96]), indicating a very high consistency rate between the two individuals' judgments and a high level of credibility in the results.

## About the focus points

What we are most concerned about is undoubtedly the factor of "focus points," because the results presented in this article do not seem to be completely consistent with the ideas advocated by international science education at present.

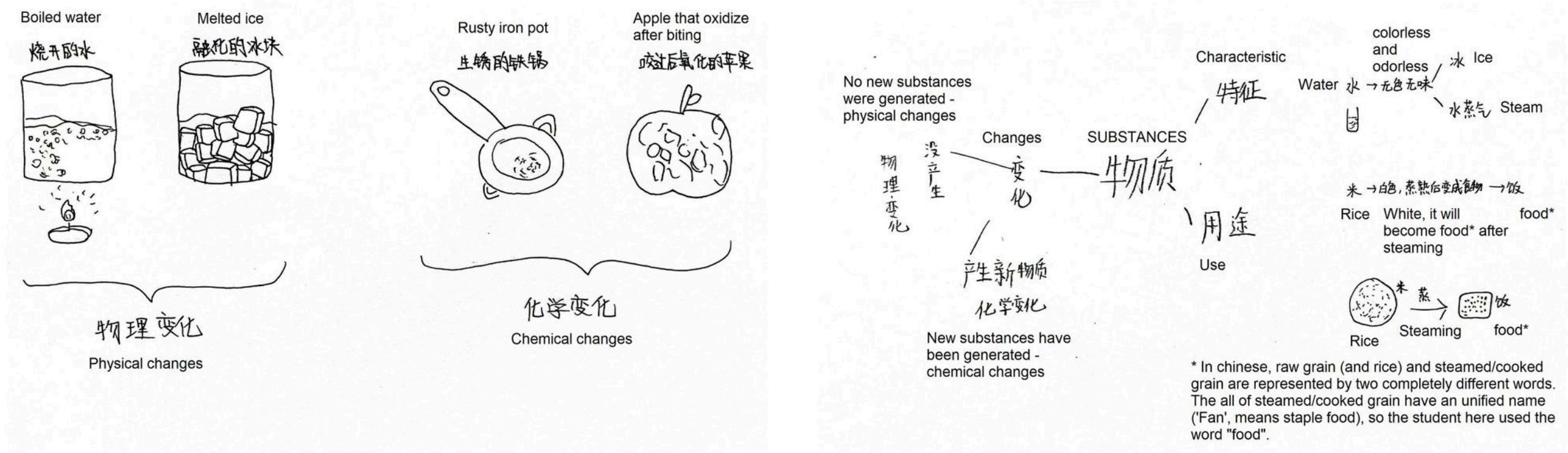

**Figure 27** Left: using pictures to represent this abstract theme, and dividing the pictures into two groups: physical changes (left) and chemical changes (right); right: using mind maps to represent this abstract theme

Undoubtedly, in the field of science education today, it is the mainstream to advocate teaching more implicit abilities such as reasoning and critical thinking in science classrooms, rather than just allowing students to memorize and apply concrete knowledge. Both Bloom's theory [28] and internationally renowned literature on science education such as NGSS [29] or PISA [30] place great emphasis on these implicit abilities. In other words, the lessons advocated by science education should not only focus on knowledge and inquiry activities themselves, but should also encourage students to infer the true results from these specific activities—i.e. "what these experimental results indicate/prove."

However, from Piaget's perspective, can school-age children really handle so many learning tasks that contain implicit components? No later than 1962, Jean Piaget published his famous theory of cognitive development stages in children [31]. In this theory, Piaget classified the cognitive development of children aged 7–12 years into concrete operations stages (COS). Piaget believed that although children in this stage have more mature abilities such as classification and decentralization compared to the previous stage (pre-school), their thinking is still limited to concrete things that are real or imagined, and they lack the ability to use abstract ways of thinking [20]. Although this theory has received some questions [32], it remains one of the most important theories in developmental psychology. And this theory is also supported by some neuroscience evidence, such as the developmental process of default networks [33, 34]. Especially H.M. Dong *et al.* [35] proposed that there is a sharp turning point in the development process of this network around the age of 13 years, which promotes the transition of cognitive processing from concrete based to abstract based]. And at the age of 13 years, it happens to be very close to the end point of the COS stage.

Our results have almost fully validated this theory. As shown in the children's drawing of the "Boiling" concept, compared to the changes in the state of water during heating, children are obviously more concerned about experimental instruments such as beakers and alcohol lamps. This is not difficult to understand, and experimental equipment is more concrete and easily accessible than state changes. And "Increasing the carrying capacity of a boat" is more typical and interesting—the children seem to have completely forgotten that this lesson should have had two items, and instead focused all their attention on the more concrete content of making a small boat, rather than adjusting the variable (carrying capacity). Their brains seem to have unintentionally "declined" this implicit task.

However, we cannot completely avoid abstract and implicit learning tasks in science class, so what should we do? A simple approach is to emphasize learning goals, especially implicit goals, to students before learning. Another approach is to modify the teaching process, reducing concrete content and emphasizing abstract content more (such as directly providing a small boat for students and using most of the class time to adjust the load capacity). But when we sought advice from frontline teachers, we received a more interesting suggestion: try to concretize abstract tasks. Many teaching methods can help achieve this idea, such as STEAM [36]. In addition, the development of computer technology has given this idea more possibilities, such as using visualization methods to visualize abstract mathematical solving processes [37]. Especially with the development of artificial intelligence technology (especially LLM), learners can have timely conversations with LLM in the classroom and receive timely feedback through text and images, which helps to visualize abstract tasks [38]. The exploration of above has received extensive teaching and academic discussions in China [39].

Of course, due to the limitations of researchers' own abilities, the number of drawings we can process is not yet sufficient to establish a norm that covers all themes/concepts in science education. But we believe that the above results can provide some reference value for children's drawing research and reveal some cognitive rules of children toward scientific concepts. We also hope that more researchers can further develop the methods in this article to analyze more children's drawings, in order to truly establish a norm that covers all themes/concepts.

In addition, there are three shortcomings in this study: firstly, all student participants are from Beijing, China, and most participants use the same version of textbooks, which undoubtedly limits the universality of the conclusions of this study in different educational and cross-cultural backgrounds. Therefore, future researches will strive to make up for this deficiency by including participants from different regions and educational environments (e.g. students in underdeveloped areas and who use different versions of textbooks), in order to increase the universality of research conclusions. Secondly, The high *P*-value and low ICC (3,1) value indicate of the Likert scale that we have not actually found any relevant factors for the representation of children's drawings in abstract concepts, suggesting that this issue may be more complex than we imagine. Given that this section is not the main focus of this article and we are not experts in studying abstract concepts, this question can only be left to other interested researchers to answer. Thirdly, due to limited manpower and energy of the research team, we only examined nine concepts/

themes. This is also the main reason why we use Kendall's τ for correlation coefficient calculation. However, such a small sample size inevitably reduces the reliability of the results. Therefore, it is recommended that readers consider the results in the "Analysis of related factors" section of this article as a reference only.

## Conclusion

This article provides a novel approach to understanding children's drawings, which utilizes LLM to convert drawing information into more accurate linguistic information, allowing for in-depth research on children's conceptual learning using linguistic tools (such as NLP algorithm). The experimental results indicate that, the representation of most drawings has consistency, manifested as most semantic similarity >0.8. And it was found that the consistency of the representation is independent of the accuracy (of LLM's recognition), indicating the existence of consistency bias. The important reason for this phenomenon is the bias between classroom experiments and teaching purpose, many students focus more on the experiments themselves rather than what they explain. In addition, most children tend to use examples they have seen in class to represent more abstract themes/concepts, indicating that they may need concrete examples to understand abstract things.

## Acknowledgments

We would like to give special thanks to Jinlong WANG, the science teacher of Nancai Town No.2 Primary School, Shunyi, Beijing. He provided valuable inspiration in the "Discussion" section (the paragraph next Piaget's Theory) of this article on how to improve teaching the abstract and implicit learning tasks.

## Author contributions

Yi Zhang (Data curation [lead], Funding acquisition [lead], Project administration [lead], Resources [lead], Writing—review & editing [lead]), Fan Wei (Data curation [lead], Formal analysis [supporting], Investigation [supporting], Methodology [supporting], Writing—original draft [supporting]), Jingyi Li (Conceptualization [supporting], Data curation [supporting], Formal analysis [supporting], Investigation [supporting], Software [supporting], Validation [supporting], Writing—original draft [lead]), Yan Wang (Investigation [supporting], Validation [supporting], Visualization [supporting]), Yanyan Yu (Investigation [supporting], Validation [supporting]), Jianli Chen (Investigation [supporting], Validation [supporting]), Zipo Cai (Investigation [supporting], Validation [supporting]), Xinyu Liu (Investigation [supporting], Validation [supporting]), Wei Wang (Investigation [supporting], Validation [supporting]), Peng Wang (Investigation [supporting], Validation [supporting]), and Sensen Yao (Methodology [supporting], Resources [supporting])

*Conflict of interest statement*. None declared.

## Funding

This work was supported by Beijing Education Society (FTYB2021-077).

## Data availability

All experimental data have been uploaded to: https://doi.org/10.57760/sciencedb.10928. Due to the involvement of children's personal information, if need to view the original drawings, please contact the corresponding author of this article to obtain electronic photos of the original drawings. The pseudocode of algorithm of word2vec and TF-IDF has been uploaded to the webpage (it can be publicly accessed): https://www.researchgate.net/publication/392771275_Pseudocodepdf.